\documentclass{article}

\usepackage{microtype}
\usepackage{graphicx}
\usepackage{subfigure}
\usepackage{bm}
\usepackage{booktabs} %

\usepackage{hyperref}

\usepackage[table]{xcolor}

\usepackage{caption}
\usepackage[accepted]{icml2025}

\usepackage{amsmath}
\usepackage{amssymb}
\usepackage{mathtools}
\usepackage{amsthm}
\usepackage{booktabs}
\usepackage{multirow}
\usepackage{array}
\usepackage[table]{xcolor}
\usepackage[capitalize,noabbrev]{cleveref}
\usepackage{enumitem}
\usepackage{pifont}
\newcommand{\xmark}{\ding{55}}
\theoremstyle{plain}
\newtheorem{theorem}{Theorem}[section]
\newtheorem{proposition}[theorem]{Proposition}

\newtheorem{corollary}[theorem]{Corollary}
\theoremstyle{definition}

\theoremstyle{remark}

\usepackage{xcolor}

\usepackage{array,booktabs,multirow,makecell,tabularx}
\usepackage[table]{xcolor}
\usepackage[textsize=tiny]{todonotes}

\usepackage{array,booktabs,graphicx,makecell,multirow}
\usepackage[table]{xcolor}
\usepackage{amsmath,amssymb,bm,booktabs,graphicx}

\newcommand{\model}{\textbf{KDA+}}
\icmltitlerunning{\model}

\newcommand{\introductionepigraph}{%
  \par\nobreak\noindent\hfill
  \parbox{0.68\linewidth}{%
    \normalfont\fontsize{9}{11}\selectfont\itshape
    \raggedleft
    ``A small leak will sink a great ship.''\par
    \vspace{3pt}\hrule height 0.3pt\relax\vspace{3pt}
    -- Benjamin Franklin\par
  }%
  \par\nobreak
}

\begin{document}

\twocolumn[{

\icmltitle{KDA+: Understanding and Mitigating Recurrent State Instability}

\begin{center}
\begin{minipage}{\textwidth}
\centering
\normalfont
{\fontsize{11}{14}\selectfont
Xichong Zhang\textsuperscript{1*}\quad
Xixuan Hao\textsuperscript{2*}\quad
Zhichen Liu\textsuperscript{3*}\quad
Le Zhou\textsuperscript{4*}\quad\\[3pt]
Zeyu Zhang\textsuperscript{5}\quad
Boyu Chen\textsuperscript{4}\quad
Mingfei Lu\textsuperscript{4}\quad
Zehao Lin\textsuperscript{4}\quad
Bo Tang\textsuperscript{1}\quad\\[3pt]
Yonggang Zhang\textsuperscript{6\ensuremath{\dagger}}\quad
Linyun Liu\textsuperscript{7\ensuremath{\dagger}}\quad
Feiyu Xiong\textsuperscript{4}\quad
Jianlin Su\textsuperscript{8}\quad
Zhiyu Li\textsuperscript{4}\par}

\vspace{10pt}
{\fontsize{9.5}{12}\selectfont
\textsuperscript{1}USTC\quad
\textsuperscript{2}HKUST(GZ)\quad
\textsuperscript{3}SUSTech\quad
\textsuperscript{4}SJTU\quad
\textsuperscript{5}RUC\quad
\textsuperscript{6}HKUST\quad
\textsuperscript{7}CQU\quad
\textsuperscript{8}kexue.fm\par}

\vspace{5pt}
{%
\textsuperscript{*}Equal contribution.\quad
\textsuperscript{\ensuremath{\dagger}}Corresponding authors.
\par}

\end{minipage}
\end{center}
\vspace{10pt}
}]

\hypersetup{pdfauthor={Xichong Zhang, Xixuan Hao, Zhichen Liu, Le Zhou, Zeyu Zhang, Boyu Chen, Mingfei Lu, Zehao Lin, Bo Tang, Yonggang Zhang, Linyun Liu, Feiyu Xiong, Jianlin Su, Zhiyu Li}}
\icmlkeywords{Machine Learning, ICML}

\begingroup
\makeatletter
\renewcommand{\@makefntext}[1]{\noindent#1}
\makeatother
\footnotetext{%
  \raggedright
  Contact: \\
  Xichong Zhang \textless yyu18@mail.ustc.edu.cn\textgreater, \\
  Xixuan Hao \textless xhao390@connect.hkust-gz.edu.cn\textgreater, \\
  Zhichen Liu \textless liuzc2024@mail.sustech.edu.cn\textgreater, \\
  Le Zhou \textless zhoule116@sjtu.edu.cn\textgreater, \\
  Yonggang Zhang \textless zhangyg@ust.hk\textgreater, \\
  Linyun Liu \textless liulinyun.oo@gmail.com\textgreater, \\
  Jianlin Su \textless bojone@spaces.ac.cn\textgreater.\par
  Zhiyu Li \textless zhiyulee@icloud.com\textgreater.\par
}
\endgroup

\begin{abstract}

Linear attention offers an efficient solution for long-sequence modeling via linear complexity and fixed-size recurrent states. Gated DeltaNet-2 (GDN-2) introduces independent channel-wise control over erasure and writing, enhancing state-update expressiveness and in-context retrieval, yet it suffers from training failures during long-context training. We trace these failures to the asymmetry between reading and modification directions during erasure, which yields non-normal erase operators whose directional gains accumulate across tokens, triggering exponential state growth even without new writes. 
To resolve this, we draw on the symmetric delta update of Kimi Delta Attention (KDA) to develop \model{}, which introduces independent channel-wise erase and write gates while preserving non-expansive state propagation and uses decay-regulated residual writing to preserve flexibility.
Theoretically, we establish an $O(\sqrt{T})$ upper bound on recurrent state growth while retaining chunkwise parallel computation. Experiments on 1.6B parameter models show that \model{} achieves stable training alongside improved long-context retrieval. Compared to GDN-2, it boosts the 128K average RULER score by 6.80 points in the recurrent setting and average accuracy across nine downstream tasks by 0.62 points in the hybrid setting.

\end{abstract}

\section{Introduction}
\introductionepigraph

The growing demand for long-context understanding and extended generation~\cite{sun2025efficient,zhang2026survey} has made efficient sequence modeling an increasingly important challenge for large language models. 
To address the computational and memory bottlenecks of standard softmax attention~\cite{llama2023llama,lu2025moba}, linear attention~\cite{katharopoulos2020transformers,retnet} has emerged as an efficient alternative that compresses historical information into a fixed-size recurrent state, thereby enabling linear-time sequence processing and constant memory during autoregressive decoding. Building on this approach, recent advances in gating mechanisms and delta-rule-based updates~\cite{gla,GDN,yang2024parallelizing} have substantially enhanced the expressiveness of recurrent memory, enabling more fine-grained control over information retention and modification.
Gated DeltaNet-2 (GDN-2)~\cite{GDN2} further advances this line of work by replacing the shared scalar update gate in Gated DeltaNet (GDN)~\cite{GDN} and Kimi Delta Attention (KDA)~\cite{kda} with independent channel-wise erase and write gates, supporting more expressive recurrent state updates and improved in-context retrieval. Specifically, the erase gate reweights the key coordinates used to read content for erasure, while the modification is applied along the original key, allowing more flexible selection of content for erasure.

\begin{figure}[]
    \centering
\includegraphics[width=0.99\linewidth]{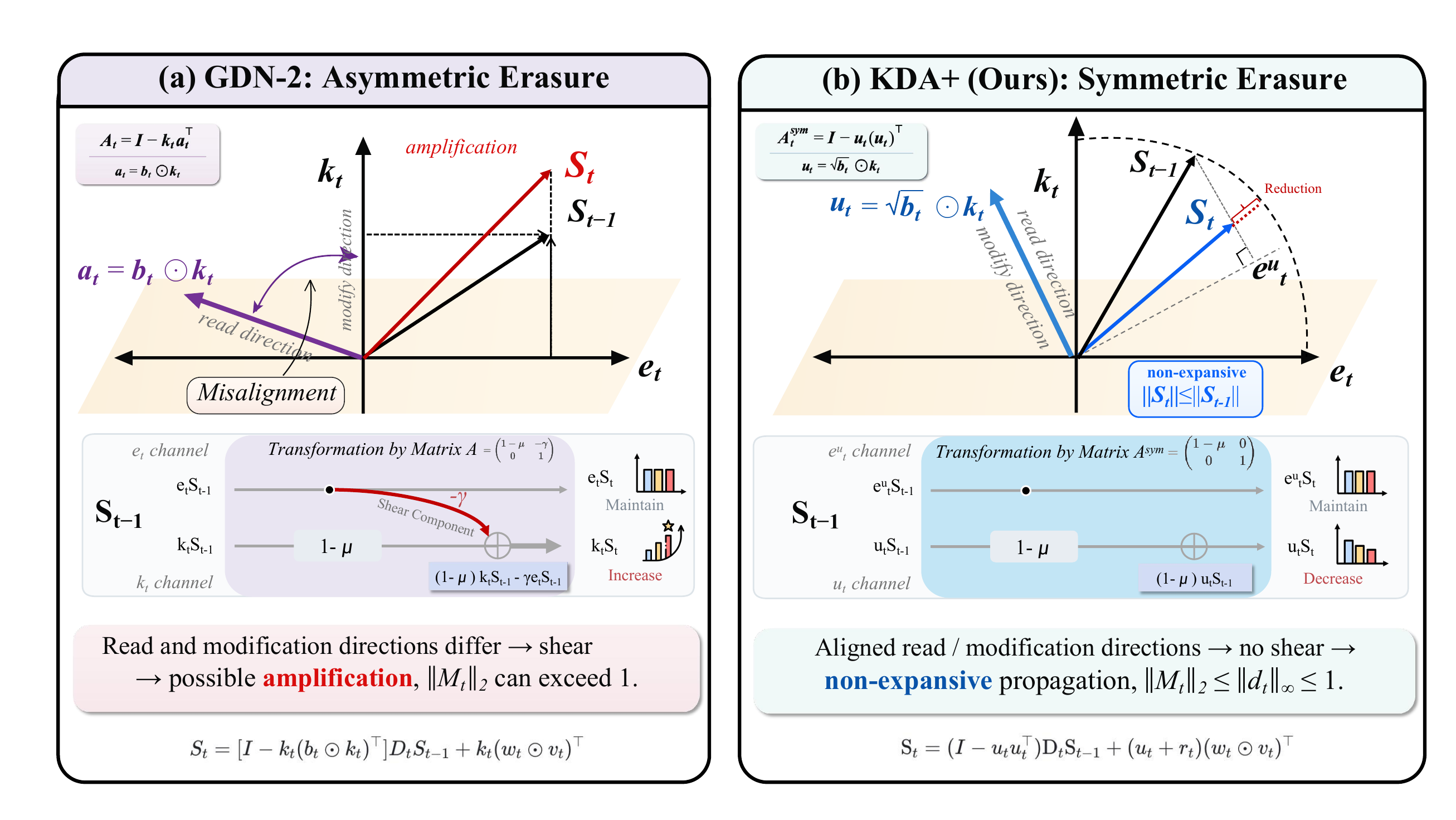}
    \caption{Geometric comparison between GDN-2 and \model{}.}
    \label{fig:intro}
    \vspace{-1em}
\end{figure}

However, we argue that \textbf{the asymmetry between the reading and modification directions during erasure in GDN-2 constitutes an inherent source of \textit{instability} in its recurrent dynamics}, potentially triggering catastrophic state explosion and ultimately causing training failures at long context.
Our analysis shows that the asymmetric coupling between the erase gate and the key can introduce shear, preserving some state components while inducing additional components along other directions, as illustrated in Figure~\ref{fig:intro}(a).
The resulting non-normal erasure operator can amplify certain state directions despite all its eigenvalues having magnitude at most one.
When amplified output directions of earlier transitions align with expanding input directions of subsequent transitions, these gains can accumulate across the sequence, allowing the recurrent state to continue growing to the point of numerical overflow, even after new writes cease.
As Franklin famously remarked, “A small leak will sink a great ship.”

To thoroughly investigate the underlying mechanisms of this instability, we examine recurrent state dynamics and their numerical and representational consequences in practical training.
At a context length of 64K, we observe numerical failures across both pretraining and post-training settings, affecting recurrent as well as hybrid GDN-2 architectures. More importantly, numerical failures represent only the most visible manifestation of a broader issue: even in runs that complete training with finite losses, individual entries of the recurrent state can exceed \(10^{36}\), indicating substantial internal state amplification. 
Replay analysis shows that amplification from asymmetric erasure can concentrate state energy in a dominant direction and sharply reduce effective capacity before numerical failure, with little warning in the training loss.
These observations highlight a fundamental limitation of loss-based monitoring and motivate a closer investigation of the underlying recurrent state dynamics that govern memory propagation over long sequences.

These findings motivate building on the symmetric delta update of KDA, which combines channel-wise decay with non-expansive erasure but couples erasure and writing through a shared scalar gate.
We therefore propose \model{}, which extends this update with independent channel-wise erase and write gates.
Distributing the erase gate symmetrically across the two key factors aligns the reading and modification directions, ensuring non-expansive propagation under the standard key normalization and gate constraints, as illustrated in Figure~\ref{fig:intro}(b). To complement the resulting gated write, we introduce a residual writing path whose amplitude is regulated by decay. This design supports additional writing in channels where the erase gate is small while controlling the energy introduced by the residual. The update retains an input-dependent affine structure compatible with chunk-wise parallel computation.

We evaluate \model{} through 64K-context pretraining in recurrent and hybrid architectures, comparing language modeling, downstream performance, and long-context retrieval against KDA, GDN-2, and existing stabilization methods.
The results show that controlling recurrent state amplification while retaining channelwise erase gating supports improved retrieval and downstream performance, alongside competitive language modeling quality.

In summary, our contributions lie in the following aspects:
\begin{itemize}[leftmargin=*,topsep=0.1em]
    \item \textit{Understanding recurrent state instability in GDN-2.}
    We characterize recurrent state instability in GDN-2 theoretically and empirically, showing that amplification can persist even without further writes.
    To the best of our knowledge, this is the first work to empirically identify training instability in GDN-2 and trace its origin to asymmetric erasure.
    \item \textit{A principled stabilization mechanism for GDN-2.}
    We propose \model, which employs symmetric erasure to guarantee non-expansive state propagation while retaining channel-wise gating. We further introduce a decay-regulated residual writing path to preserve writing flexibility without compromising the stability guarantee. The resulting formulation supports chunkwise parallel computation.
    \item \textit{Comprehensive empirical evaluation.}
    We evaluate \model{} against KDA, GDN-2, and existing stabilization methods using recurrent and hybrid models pretrained on 64K-token sequences.
    Compared with GDN-2, \model{} improves the average RULER score at 128K by 6.80 points in the recurrent setting and average accuracy across nine downstream tasks by 0.62 percentage points in the hybrid setting, while maintaining competitive language modeling performance.
    
\end{itemize}

\section{Preliminary}
\label{sec:preliminaries}

We consider a single attention head and omit head and layer indices.
At position $t$, the query and key are $\bm{q}_t,\bm{k}_t\in\mathbb{R}^{d_k}$, the value is $\bm{v}_t\in\mathbb{R}^{d_v}$, and the recurrent state is $\mathbf{S}_t\in\mathbb{R}^{d_k\times d_v}$.
All vectors are column vectors.
We use $\odot$ for elementwise multiplication and $\operatorname{Diag}(\bm{x})$ for the diagonal matrix with diagonal $\bm{x}$.

\subsection{Linear Attention and the Gated Delta Rule}
\label{sec:prelim:delta_rule}

Linear attention accumulates historical key--value associations in a fixed-size matrix state and retrieves them using the current query~\cite{katharopoulos2020transformers}.
Following recent delta-rule models~\cite{yang2024parallelizing,GDN}, we adopt the linear-kernel formulation without an attention normalizer:
\begin{equation}
    \mathbf{S}_t=\mathbf{S}_{t-1}+\bm{k}_t\bm{v}_t^\top,
    \quad
    \bm{o}_t=
    \mathbf{S}_t^\top\bm{q}_t.
    \label{eq:linear_attention}
\end{equation}
Each outer product adds a new association to the memory, whose dimensions are independent of sequence length.
This additive update does not explicitly revise the content already associated with the incoming key.

The delta rule corrects an existing association using the residual between the incoming value and the value retrieved from memory~\cite{schlag2021linear,yang2024parallelizing}.
For a scalar update gate $\beta_t\in[0,1]$, DeltaNet applies
\begin{equation}
\begin{aligned}
    \mathbf{S}_t &=\mathbf{S}_{t-1}+\beta_t\bm{k}_t
    \bigl(\bm{v}_t-\mathbf{S}_{t-1}^\top\bm{k}_t\bigr)^\top\\
    &=  \bigl(\mathbf{I}-\beta_t\bm{k}_t\bm{k}_t^\top\bigr)
    \mathbf{S}_{t-1}+  \beta_t\bm{k}_t\bm{v}_t^\top.
\end{aligned}
\label{eq:delta_rule}
\end{equation}
Here, $\mathbf{S}_{t-1}^\top\bm{k}_t$ is the value predicted by the current memory.
We consider $\ell_2$-normalized keys, $\|\bm{k}_t\|_2=1$, throughout the delta-rule recurrences below.
The expanded form separates the removal of retrieved content from the insertion of the new value.
Both operations are controlled by the same scalar gate $\beta_t$.
Gated DeltaNet introduces a forget gate that decays the existing memory before applying the delta-rule correction~\cite{GDN}.
The resulting update takes the form
\begin{equation}
    \mathbf{S}_t
    =
    \bigl(\mathbf{I}-\beta_t\bm{k}_t\bm{k}_t^\top\bigr)
    \mathbf{D}_t\mathbf{S}_{t-1}
    +
    \beta_t\bm{k}_t\bm{v}_t^\top.
    \label{eq:gated_delta_rule}
\end{equation}
Gated DeltaNet uses a scalar decay, $\mathbf D_t=d_t\mathbf I$ with $d_t\in(0,1]$.
Kimi Delta Attention instead uses $\mathbf D_t=\operatorname{Diag}(\mathbf d_t)$, where $\mathbf d_t\in(0,1]^{d_k}$ contains a separate decay factor for each key channel~\cite{kda}.
Setting $\mathbf{D}_t=\mathbf{I}$ recovers DeltaNet.
Although KDA provides finer control over forgetting, its erase and write operations remain tied to the same scalar update gate.

\subsection{Gated DeltaNet-2}
GDN-2 replaces the shared scalar update gate with separate channel-wise gates for erasing and writing~\cite{GDN2}.
Let $\bm{b}_t\in(0,1)^{d_k}$ and $\bm{w}_t\in(0,1)^{d_v}$ denote the erase and write gates, respectively.

Retaining KDA's diagonal decay, GDN-2 updates the state as
\begin{equation}
\begin{aligned}
    \mathbf{S}_t
    ={}&
    \bigl[
        \mathbf{I} - \bm{k}_t(\bm{b}_t\odot\bm{k}_t)^\top
    \bigr]
    \mathbf{D}_t\mathbf{S}_{t-1}
    +
    \bm{k}_t(\bm{w}_t\odot\bm{v}_t)^\top.
\end{aligned}
\label{eq:GDN-2}
\end{equation}
The erase gate weights the key coordinates used to retrieve old content for removal, while the write gate scales the value coordinates committed to memory.
These gates are input-dependent and parameterized independently.

GDN-2 allows the erase operation to read and modify the state along different directions.
In GDN and KDA, both the retrieval of old content and the subsequent modification use the key $\bm{k}_t$.
In GDN-2, retrieval uses $\bm{b}_t\odot\bm{k}_t$, while the modification is still applied along $\bm{k}_t$.
These vectors need not be parallel, so the erase matrix $\mathbf{I}-\bm{k}_t(\bm{b}_t\odot\bm{k}_t)^\top$ need not be symmetric.
This structural change arises from the channel-wise erase gate; the independent write gate determines the content injected into the state.
The recurrence reduces to KDA when $\bm{b}_t=\beta_t\mathbf{1}_{d_k}$ and $\bm{w}_t=\beta_t\mathbf{1}_{d_v}$, and further reduces to Gated DeltaNet when $\mathbf D_t=d_t\mathbf I$.

GDN-2 evaluates the same recurrence through chunkwise parallel computation during training~\citep{GDN2}.
For chunks of length $C$, let $\mathbf{S}_{[n]}=\mathbf{S}_{nC}$ denote the state entering chunk $n$, which contains positions $nC+1,\ldots,(n+1)C$; its exit state is $\mathbf{S}_{[n+1]}=\mathbf{S}_{(n+1)C}$.
Token interactions within each chunk are computed through parallel matrix operations, while the boundary states carry information between successive chunks.

\section{Understanding State Instability in GDN-2}
\label{sec:instability}

In GDN-2, the asymmetric coupling between the channel-wise erase gate and the key yields a generally non-normal erase operator, breaking the non-expansiveness guaranteed by symmetric erasure in GDN and KDA. Across time-varying transitions, directional gains from this non-normality can accumulate when amplified state directions align with subsequent expanding directions, allowing stored memory to grow even after new writes cease. In the following, we theoretically formalize this mechanism (Section~\ref{sec:theoretical_analysis}) and show that it manifests in long-context training as both sudden numerical failures and extreme state excursions hidden beneath normal loss trajectories (Section~\ref{sec:empirical_characterization}).

\subsection{Theoretical Analysis}
\label{sec:theoretical_analysis}

GDN-2 employs channel-wise erase gates to offer fine-grained control over recurrent states.
Erasure reads the decayed recurrent state $\mathbf{D}_t\mathbf{S}_{t-1}$ using the gated key $\mathbf{b}_t\odot\mathbf{k}_t$ and applies the resulting modification along the original key $\mathbf{k}_t$.
The stability question is whether this asymmetric modification can amplify the recurrent state even when every gate entry lies in $(0,1)$.

The nontrivial action of erasure is confined to the subspace spanned by the key and its gated counterpart.
Omitting the time index, let $\mathbf{a}=\mathbf{b}\odot\mathbf{k}$ and
$\mathbf{A}=\mathbf{I}-\mathbf{k}\mathbf{a}^{\top}$.
For $\|\mathbf{k}\|_2=1$, choose $\mathbf{e}_{\parallel}=\mathbf{k}$ and decompose
\begin{equation}
    \begin{aligned}
        \mathbf{a}&=\mu\mathbf{e}_{\parallel}+\gamma\mathbf{e}_{\perp},\\
        \mu&=\mathbf{k}^{\top}\mathbf{a},
        \qquad\gamma=\|\mathbf{a}-\mu\mathbf{k}\|_2,
    \end{aligned}
    \label{eq:gated_key_decomposition}
\end{equation}
where $\mathbf{e}_{\perp}$ is a unit vector orthogonal to
$\mathbf{e}_{\parallel}$; when $\gamma=0$, any such vector may be chosen.
Expressing $\mathbf{A}\mathbf{e}_{\parallel}=(1-\mu)\mathbf{e}_{\parallel}$
and $\mathbf{A}\mathbf{e}_{\perp}=\mathbf{e}_{\perp}-\gamma\mathbf{e}_{\parallel}$
in this orthonormal basis gives the two-dimensional representation
\begin{equation}
    \mathbf{T}=
    \begin{pmatrix}
        1-\mu & -\gamma \\
        0 & 1
    \end{pmatrix},
    \label{eq:GDN-2_erase_geometry}
\end{equation}
with the identity acting on the orthogonal complement.
Here, $\mu$ controls attenuation along the key, while $\gamma$ measures the shear. It means that 
an orthogonal component is retained while contributing an additional component along the key.
This coupling depends jointly on the erase gate and the key.

Nonzero shear $\gamma$ creates directions in which erasure amplifies the recurrent state. The following theorem gives the exact maximum gain of a single erasure step.
\begin{theorem}[Single-step amplification in GDN-2]
\label{thm:GDN-2_single_step}
Let $\mathbf{A}=\mathbf{I}-\mathbf{k}(\mathbf{b}\odot\mathbf{k})^{\top}$,
where $d_k\geq2$, $\|\mathbf{k}\|_2=1$, and
$\mathbf{b}\in[0,1]^{d_k}$.
With $\mu$ and $\gamma$ defined in Eq.~\eqref{eq:gated_key_decomposition},
its operator spectral norm is
\begin{equation}
    \begin{aligned}
        \|\mathbf{A}\|_2&=\sigma_{\max}(\mathbf{A})\\
        &=\frac{\sqrt{\mu^2+\gamma^2}
        +\sqrt{(2-\mu)^2+\gamma^2}}{2}.
    \end{aligned}
    \label{eq:GDN-2_erasure_norm}
\end{equation}
In particular, $\|\mathbf{A}\|_2=1$ if and only if $\gamma=0$,
equivalently, $\mathbf{a}$ and $\mathbf{k}$ are collinear.
Otherwise, $\|\mathbf{A}\|_2>1$.
\end{theorem}
The proof is given in Appendix~\ref{app:GDN-2_single_step}. The spectral norm $\|\mathbf A\|_2$ measures the maximum amplification factor of the recurrent state norm, $\max_{\mathbf S\neq\mathbf0}\|\mathbf A\mathbf S\|_F/\|\mathbf S\|_F$.
Accordingly, Theorem~\ref{thm:GDN-2_single_step} shows that erasure alone can increase the energy of recurrent state,
defined as $E(\mathbf{S})=\|\mathbf{S}\|_F^2$, even without any new write.
For GDN and KDA, scalar erase gating gives $\mathbf{a}=\beta\mathbf{k}$
and $\gamma=0$, recovering non-expansive symmetric erasure.
As a simple example of the geometry in Figure~\ref{fig:intro},
consider a single-column recurrent state $\mathbf{S}_t=\mathbf{e}_{\perp}$.
When $\gamma>0$, erasure gives
\begin{equation}
    \begin{aligned}
        \mathbf{A}\mathbf{S}_t&=\mathbf{e}_{\perp}-\gamma\mathbf{e}_{\parallel},\\
        E(\mathbf{A}\mathbf{S}_t)&=1+\gamma^2>E(\mathbf{S}_t)=1.
    \end{aligned}
    \label{eq:GDN-2_orthogonal_example}
\end{equation}
This amplification can persist in the full transition
$\mathbf{M}=\mathbf{A}\mathbf{D}$.
In particular, an admissible scalar decay $\mathbf{D}=d\mathbf{I}$
with $1/\|\mathbf{A}\|_2<d<1$ gives
$\|\mathbf{M}\|_2=d\|\mathbf{A}\|_2>1$.

The analysis of a single update shows that a full transition $\mathbf{M}_t=\mathbf{A}_t\mathbf{D}_t$ can increase recurrent state
energy even without a new write.
Along a token sequence, the recurrent state evolves through successive
transitions that depend on the input. Their composition determines whether this amplification can persist.
Suppose no further content is written after token $s$.
The recurrent state at token $t>s$ then satisfies
\begin{equation}
    \begin{aligned}
        \mathbf{S}_t
        &=\boldsymbol{\Phi}_{t:s+1}\mathbf{S}_s,
    \end{aligned}
    \label{eq:GDN-2_state_expansion}
\end{equation}
 where $\boldsymbol{\Phi}_{t:s+1} =\mathbf{M}_t\cdots\mathbf{M}_{s+1}$.
When an amplified output direction of $\mathbf{M}_j$ aligns with an expanding input direction of $\mathbf{M}_{j+1}$, the transition at the next token can amplify it further.
Repeated alignment across tokens can therefore allow gains to accumulate, posing a risk of sustained recurrent state energy growth even after writing has ceased.

Such alignment can be realized by an admissible sequence of GDN-2 updates,
yielding unbounded recurrent state growth after a single finite write.
\begin{corollary}[Unbounded growth without further writes]
\label{cor:GDN-2_state_growth}
There exist normalized keys, together with erase, write, and decay gates
whose entries lie in $(0,1)$, such that the GDN-2 recurrence,
initialized at $\mathbf{S}_0=\mathbf{0}$, satisfies
\begin{equation}
    \lim_{t\to\infty}\|\mathbf{S}_t\|_F=\infty,
    \label{eq:GDN-2_unbounded_growth}
\end{equation}
after a finite nonzero write at the first token, even with no writes at subsequent tokens.
\end{corollary}
Appendix~\ref{app:GDN-2_growth_construction} provides an explicit example
that alternates two admissible transitions with
$\mathbf{M}_b=\mathbf{M}_a^{\top}$ and
$\lambda=\sigma_{\max}(\mathbf{M}_a)^2>1$.
Starting from an initial write along a leading right singular direction
of $\mathbf{M}_a$, each pair returns the recurrent state to the same
direction and multiplies its norm by $\lambda$, yielding exponential
growth without further writes.

In summary, the asymmetric coupling between the erase gate and the key
in GDN-2 can yield a non-normal erase operator with expanding directions,
breaking the non-expansiveness of symmetric erasure in GDN and KDA. When amplified output directions align with expanding input directions
at subsequent tokens, the gains can accumulate, allowing sustained
growth in recurrent state energy without further writes and creating
a risk of numerical overflow in finite precision arithmetic.
The analysis establishes the existence of such trajectories,
while the occurrence and extent of amplification in practical training depend
on how the transitions induced by the token sequence and learned
parameters interact with the recurrent state.

\subsection{Empirical Characterization}
\label{sec:empirical_characterization}

To examine how the amplification mechanism identified above manifests in training, we study GDN-2 under pretraining, post-training, and post-training after pretraining, using Pure Linear and Hybrid architectures with four combinations of data and initialization seeds.
Pretraining uses 64K sequences to examine accumulated gains over long uninterrupted recurrent state propagation.
We record a numerical failure when the training loss becomes NaN (not a number) and prevents further training.
Across the two architectures, Table~\ref{tab:GDN-2_training_failures} reports failures in 3 of 8 pretraining runs and 4 of 8 post-training runs; all five post-training stages launched after successful pretraining complete.
Detailed evaluation protocols and failure accounting are provided in Appendix~\ref{app:empirical_protocol}.

\begin{table}[t]
    \centering
    \small
    \caption{
        Training failures across regimes and architectures.
        The designated trajectories and additional repeats are detailed in Appendix~\ref{app:empirical_protocol}.
    }
    \label{tab:GDN-2_training_failures}
    \vspace{0.5em}
    \setlength{\tabcolsep}{4pt}
    \resizebox{\columnwidth}{!}{%
        \begin{tabular}{@{}l|l|ccc@{}}
            \toprule
            \multicolumn{2}{c|}{Architecture}
            & Pretraining & Post-training
            & \makecell[c]{Pretraining $\rightarrow$\\Post-training} \\
            \midrule
            \multirow{2}{*}{GDN-2}
            & Recurrent Model & 1/4 & 1/4 & 0/3 \\
            \addlinespace[3pt]
            & Hybrid Model    & 2/4 & 3/4 & 0/2 \\
            \bottomrule
        \end{tabular}
    }
\end{table}

\begin{figure}[t]
    \centering
    \subfigure[Training Loss]{%
        \includegraphics[width=0.48\linewidth]{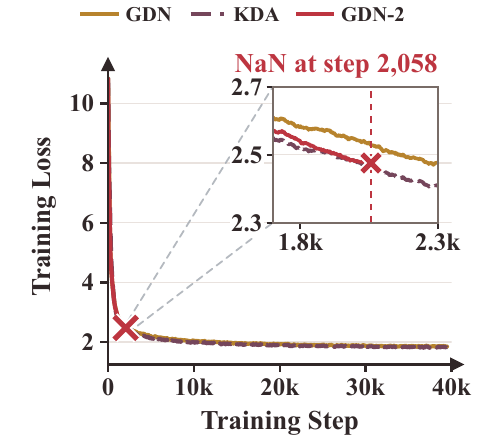}%
        \label{fig:empirical_loss}}%
    \hfill
    \subfigure[Recurrent State Norm]{%
        \includegraphics[width=0.48\linewidth]{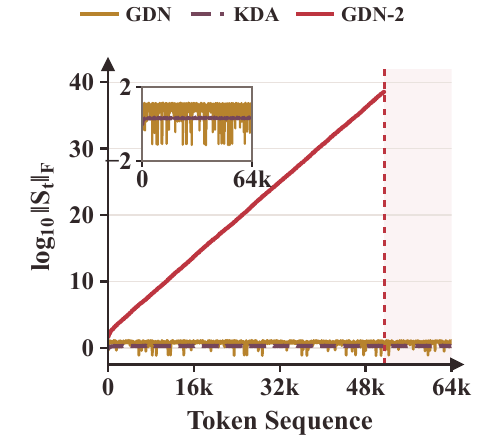}%
        \label{fig:empirical_state_norm}}%
    \\[-0.5em]
    \subfigure[Propagation without Writes]{%
        \includegraphics[width=0.48\linewidth]{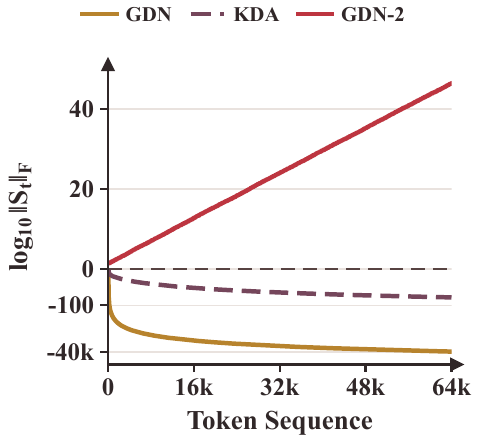}%
        \label{fig:empirical_no_write}}%
    \hfill
    \subfigure[Recurrent State Capacity]{%
        \includegraphics[width=0.48\linewidth]{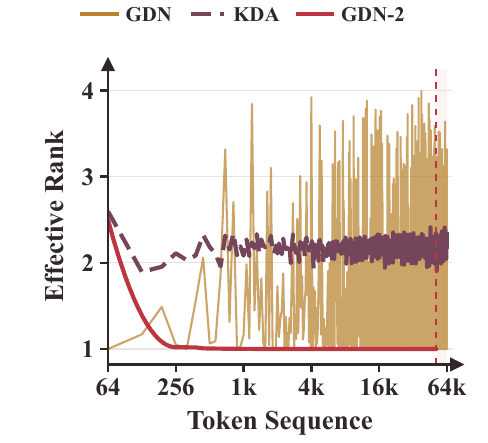}%
        \label{fig:empirical_rank}}%
    \vspace{-0.5em}
    \caption{
        The trace of a training failure.
    }
    \label{fig:GDN-2_training_diagnostics}
    \vspace{-2em}
\end{figure}

Replaying one failed pretraining run traces its loss NaN to recurrent state amplification that persists without further writes.
Figure~\ref{fig:empirical_loss} shows GDN-2 failing at step 2,058, alongside the training trajectories of GDN and KDA.
On the triggering sequence, Figure~\ref{fig:empirical_state_norm} shows sustained growth in the GDN-2 recurrent state norm, with overflow observed at token 52,736 and nonfinite values subsequently reaching the logits and loss.
GDN and KDA do not exhibit the same growth in the corresponding replay.
To isolate propagation, Figure~\ref{fig:empirical_no_write} retains the first 64 token updates and sets subsequent write terms to zero, while holding the extracted keys, gates, and decay factors fixed.
The recurrent state then follows Eq.~\eqref{eq:GDN-2_state_expansion} with $s=64$.
GDN and KDA decay after writes cease, whereas GDN-2 continues to grow.
This trajectory exhibits the mechanism established in Corollary~\ref{cor:GDN-2_state_growth} on a real token sequence.
The non-normal erase operator $A_t$ induced by the asymmetric coupling between the erase gate and the key permits expanding directions; when amplified output directions align with expanding input directions at subsequent tokens, gains can accumulate and drive recurrent state growth even without further writes.

This amplification is accompanied by a loss of effective representational dimensions before numerical overflow.
Following \citet{nazari2026state}, we measure the effective rank using $\operatorname{er}(\mathbf S_t) =\|\mathbf S_t\|_F^2/\|\mathbf S_t\|_2^2$ to characterize the effective capacity of the recurrent state.
In Figure~\ref{fig:empirical_rank}, the capacity of recurrent state using GDN-2 rapidly approaches one and remains near this value.
The concentration of energy into a single dominant direction reflects degraded utilization of the recurrent state's representational dimensions.

Extreme recurrent state magnitudes also occur during runs that eventually complete training.
Table~\ref{tab:GDN-2_seed_state_maxima} in Appendix~\ref{app:state_magnitudes} shows that the Pure Linear model with Seed~2 completes pretraining even though the largest recorded absolute recurrent state entry during training reaches $3.157\times10^{36}$.
Such values remain within the finite representable range of FP32, so substantial amplification need not immediately produce loss NaN.
These observations connect asymmetric erasure and accumulated transition gains to recurrent state growth in actual training, with consequences ranging from numerical overflow to extreme values along otherwise successful training trajectories.
They motivate directly controlling the gain of recurrent state propagation.

\section{\model{} Methodology}
\label{sec:method}

The analysis above shows that the asymmetric coupling between the channel-wise erase gate and the key in GDN-2 can amplify the recurrent state.
\textit{Can we retain the fine-grained control of erase while restoring non-expansive recurrent state propagation?}
To this end, we propose \model{}, which restores non-expansive propagation through symmetric erasure and introduces a residual write, regulated by decay, to supplement writing suppressed by symmetric gating.

\begin{figure*}[t]
    \centering
\includegraphics[width=0.99\linewidth]{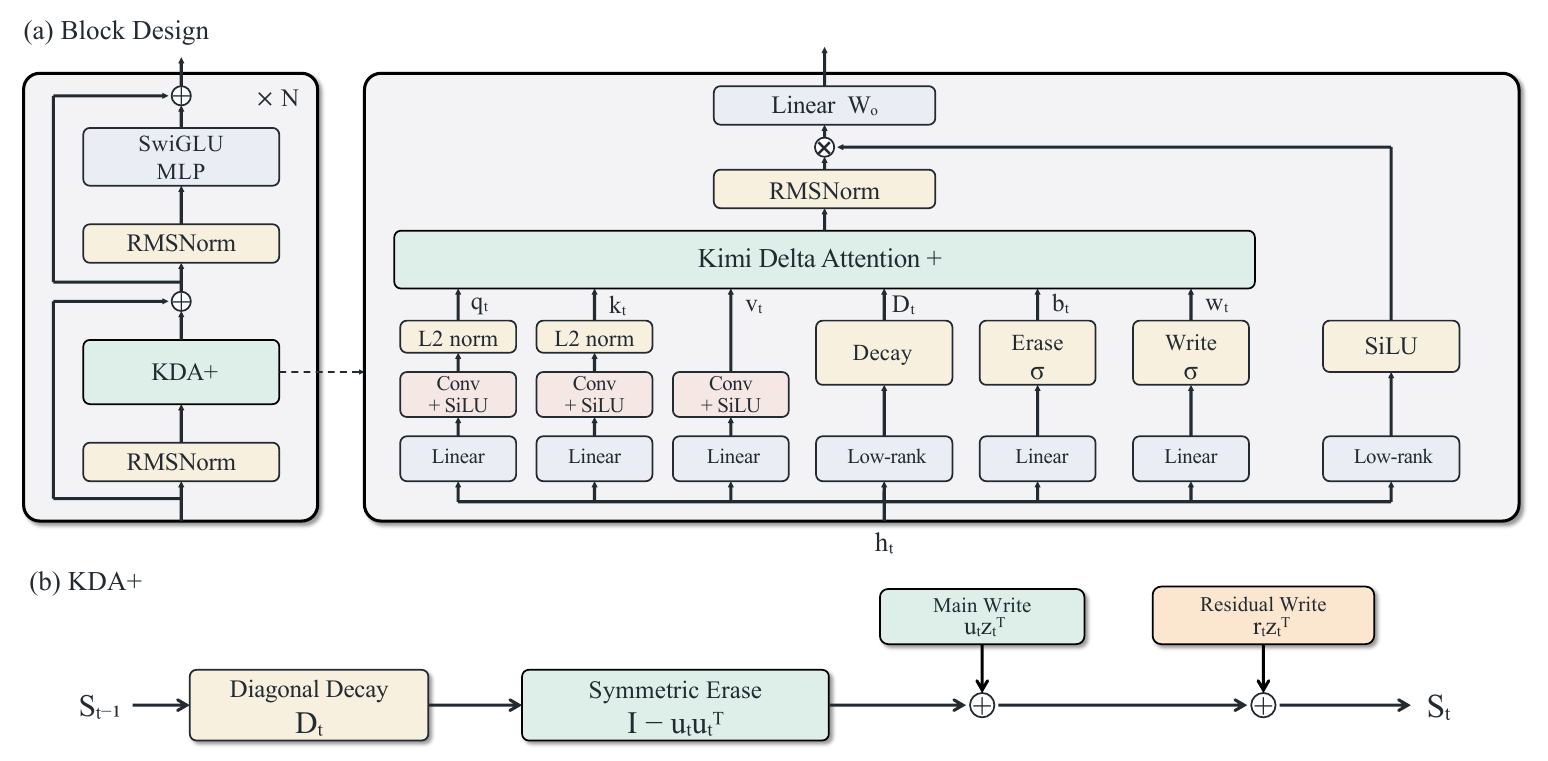}
    \caption{\model~architecture and recurrent state update.}
    \label{fig:stable_gdn2_framework}
\end{figure*}

\subsection{Symmetric Erasure and Residual Writing}
\label{sec:stable_update}

In KDA, the erase operator in Eq.~\eqref{eq:gated_delta_rule} admits the symmetric factorization $\mathbf I-\beta_t\mathbf k_t\mathbf k_t^\top
=\mathbf I-(\sqrt{\beta_t}\mathbf k_t)(\sqrt{\beta_t}\mathbf k_t)^\top$.
We extend this structure to channel-wise erase control by applying $\sqrt{\mathbf b_t}$ elementwise to both key factors, keeping the reading and modification directions aligned.
Defining $\mathbf u_t=\sqrt{\mathbf b_t}\odot\mathbf k_t$ gives the symmetric erase operator
\begin{equation}
    \begin{aligned}
        \mathbf A_t^{\mathrm{sym}}
        &=\mathbf I-
        (\sqrt{\mathbf b_t}\odot\mathbf k_t)
        (\sqrt{\mathbf b_t}\odot\mathbf k_t)^\top\\
        &=\mathbf I-\mathbf u_t
        (\mathbf u_t)^\top.
    \end{aligned}
    \label{eq:symmetric_erasure}
\end{equation}
This preserves the diagonal attenuation of the original GDN-2 operator,
$(\mathbf A_t^{\mathrm{sym}})_{ii}=(\mathbf A_t)_{ii}=1-b_{t,i}k_{t,i}^2$.

Using $\mathbf u_t$ for reading, modification, and writing, we can rewrite the delta update in this symmetric formulation as
\begin{equation}
    \mathbf S_t
    =\mathbf D_t\mathbf S_{t-1}
    +\mathbf u_t
    \left[
        \mathbf z_t-
        (\mathbf D_t\mathbf S_{t-1})^\top
        \mathbf u_t
    \right]^\top,
    \label{eq:symmetric_delta_update}
\end{equation}
where $\mathbf z_t=\mathbf w_t\odot\mathbf v_t$ is the gated value.
This retains the delta update structure and ensures non-expansive recurrent state propagation.
Equation~\eqref{eq:symmetric_delta_update} recovers the KDA recurrence in Eq.~\eqref{eq:gated_delta_rule} when
$\mathbf b_t=\beta_t\mathbf 1_{d_k}$ and $\mathbf w_t=\sqrt{\beta_t}\mathbf 1_{d_v}$.
However, as $b_{t,i}$ approaches zero, both erasure and the base write in channel $i$ vanish.
This tightly couples erasing stored content with injecting new information, restricting the ability to write while suppressing erasure.

To supplement the suppressed writes, we introduce a residual along
$\mathbf k_t-\mathbf u_t$ and regulate its amplitude by decay.
Decay retains a fraction $d_{t,i}^2$ of the recurrent state energy in channel $i$.
We therefore scale the residual by the square root of the complementary fraction,
\begin{equation}
    \mathbf r_t
    =\sqrt{1-\mathbf d_t^2}\odot
    (\mathbf k_t-\mathbf u_t).
    \label{eq:residual_write}
\end{equation}
As $d_{t,i}$ approaches one, the residual write vanishes in that channel; stronger decay permits a larger additional write.
The complete update becomes
\begin{equation}
    \begin{aligned}
        \mathbf S_t
        &=\mathbf A_t^{\mathrm{sym}}\mathbf D_t\mathbf S_{t-1}
        +
        (\mathbf u_t+\mathbf r_t)\mathbf z_t^\top.
    \end{aligned}
    \label{eq:stable_GDN-2}
\end{equation}
Beyond non-expansive propagation, this design yields a tighter bound on recurrent state energy under continued writing.
Section~\ref{sec:stability_guarantees} establishes that its growth is controlled by cumulative write energy,
$\sum_{t=1}^{T}\|\mathbf z_t\|_2^2$.

Figure~\ref{fig:stable_gdn2_framework} shows the architecture and the two write contributions.
We retain the gate parameterization of GDN-2~\citep{GDN2}, using separate sigmoid projections for the erase $\mathbf b_t$ and write gates $\mathbf w_t $ and the same exponential decay parameterization with rates constrained by softplus.
Queries and keys follow the original $\ell_2$ normalization; $\mathbf u_t$ and $\mathbf r_t$ retain their gate-dependent magnitudes without further normalization.

\subsection{Stability Guarantees}
\label{sec:stability_guarantees}

Symmetric erasure removes the expanding directions identified in Theorem~\ref{thm:GDN-2_single_step}.
For the transition $\mathbf M_t=\mathbf A_t^{\mathrm{sym}}\mathbf D_t$ in Eq.~\eqref{eq:stable_GDN-2}, this prevents recurrent state amplification in the absence of new writes.
\begin{proposition}[Non-expansive transitions without writes]
\label{prop:nonexpansive_propagation}
The symmetric erase operator in \model~satisfies
$\|\mathbf A_t^{\mathrm{sym}}\|_2\leq1$, and the corresponding transition obeys
\begin{equation}
    \|\mathbf M_t\|_2
    \leq\|\mathbf d_t\|_\infty
    \leq1.
    \label{eq:one_step_nonexpansive}
\end{equation}
When the write term at token $t$ is zero, the recurrent state therefore satisfies
$\|\mathbf S_t\|_F
\leq\|\mathbf d_t\|_\infty\|\mathbf S_{t-1}\|_F
\leq\|\mathbf S_{t-1}\|_F$.
\end{proposition}
\begin{proof}
Key normalization and the erase gate parameterization give
$\|\mathbf u_t\|_2^2
=\sum_i b_{t,i}k_{t,i}^2
\leq\|\mathbf k_t\|_2^2
\leq1$.
Hence $0\preceq\mathbf A_t^{\mathrm{sym}}\preceq\mathbf I$.
As in the analysis of symmetric erasure in GDN and KDA in Section~\ref{sec:theoretical_analysis}, the erase operator has spectral norm at most one.
By submultiplicativity,
$\|\mathbf M_t\|_2
\leq\|\mathbf A_t^{\mathrm{sym}}\|_2\|\mathbf D_t\|_2
\leq\|\mathbf d_t\|_\infty
\leq1$.
With no new write, $\mathbf S_t=\mathbf M_t\mathbf S_{t-1}$, which gives the stated Frobenius norm bound.
\end{proof}
Consequently, if writes cease after position $s$, then
$\|\mathbf S_t\|_F
\leq\|\boldsymbol\Phi_{t:s+1}\|_2\|\mathbf S_s\|_F
\leq\|\mathbf S_s\|_F$
for every $t>s$, using the propagation notation from Eq.~\eqref{eq:GDN-2_state_expansion}.
With continued writing, the recurrent state can grow through the accumulation of new content, with each contribution propagated with gain at most one.
Expanding Eq.~\eqref{eq:stable_GDN-2} gives
$\mathbf S_T
=\boldsymbol\Phi_{T:1}\mathbf S_0
+\sum_{t=1}^{T}\boldsymbol\Phi_{T:t+1}
(\mathbf u_t+\mathbf r_t)\mathbf z_t^\top$,
with empty products equal to $\mathbf I$.
Together with
$|(\mathbf u_t+\mathbf r_t)_i|\leq|k_{t,i}|$,
this yields
\begin{equation}
    \begin{aligned}
        \|\mathbf S_T\|_F
        &\leq\|\mathbf S_0\|_F+
        \sum_{t=1}^{T}
        \|(\mathbf u_t+\mathbf r_t)\mathbf z_t^\top\|_F\\
        &\leq\|\mathbf S_0\|_F+
        \sum_{t=1}^{T}\|\mathbf k_t\mathbf z_t^\top\|_F.
    \end{aligned}
    \label{eq:stable_additive_bound}
\end{equation}
Appendix~\ref{app:write_norm} gives the coordinatewise argument.
This bound has the same additive form as recurrent state accumulation in the Linear Transformer of \citet{katharopoulos2020transformers}.
For $\mathbf S_0=\mathbf0$ and $\|\mathbf v_t\|_2\leq V$, it gives
$\|\mathbf S_T\|_F\leq TV$, since
$\|\mathbf z_t\|_2
=\|\mathbf w_t\odot\mathbf v_t\|_2
\leq\|\mathbf v_t\|_2$.

Decay regulation tightens this bound from $O(T)$ to $O(\sqrt T)$ by jointly controlling propagation and writing.
\begin{proposition}[Recurrent state norm bound]
\label{prop:write_energy_bound}
Consider the \model{} recurrence in Eq.~\eqref{eq:stable_GDN-2}, initialized at $\mathbf S_0=\mathbf0$.
If $\|\mathbf v_t\|_2\leq V$ for all tokens, with a constant $V>0$, then for every $T\geq1$,
\begin{equation}
    \|\mathbf S_T\|_F
    \leq\sqrt{\frac{3T}{2}}\,V.
    \label{eq:stable_state_norm_bound}
\end{equation}
\end{proposition}
The proof in Appendix~\ref{app:input_energy} jointly bounds propagation and writing, without requiring decay factors to remain uniformly below one.
Symmetric erasure eliminates the exponential amplification through transition products identified in Section~\ref{sec:theoretical_analysis}, while decay regulation further provides an $O(\sqrt T)$ bound on the recurrent state norm under bounded values and zero initialization.
Removing decay regulation preserves non-expansive transitions but need not admit an $O(\sqrt T)$ bound with a constant independent of the gate values, as shown in Appendix~\ref{app:decay_regulation_role}.

\subsection{Chunkwise Parallel Training}
\label{sec:chunkwise_training}

Now we show that \model{} can keep the same chunkwise structure as GDN and KDA. Its transition remains Diagonal-Plus-Low-Rank (DPLR), while the residual path contributes an additional input-dependent write. Defining the modification vector $\boldsymbol{\delta}_t=\mathbf z_t-(\mathbf D_t\mathbf S_{t-1})^\top\mathbf u_t$, then Eq.~\eqref{eq:stable_GDN-2} becomes
\begin{equation}
    \mathbf S_t
    =
    \mathbf D_t\mathbf S_{t-1}
    +
    \mathbf u_t\boldsymbol{\delta}_t^\top
    +
    \mathbf r_t\mathbf z_t^\top.
    \label{eq:chunkwise_correction}
\end{equation}
Each token introduces a single state-dependent correction $\boldsymbol{\delta}_t$, while the residual write $\mathbf r_t\mathbf z_t^\top$ is fully determined by the input. The residual contributions therefore enter the correction equations as known terms on the right-hand side. Since each correction depends only on earlier corrections, these equations form a single unit lower triangular system over the token positions within a chunk.

Consider a chunk of $C$ tokens, with local indices $i=1,\ldots,C$ and initial state $\mathbf S_{\mathrm{in}}$. Following the decay factorization used in GDN-2, write $\mathbf d_i=\exp(\mathbf g_i)$ and define the cumulative decay vector $\boldsymbol{\gamma}_i =\exp\left(\sum_{j=1}^{i}\mathbf g_j\right)$, with $\boldsymbol{\gamma}_0=\mathbf 1$, where exponentials are applied elementwise. Let $\Delta,Z\in\mathbb{R}^{C\times d_v}$ contain rows $\boldsymbol{\delta}_i^\top$ and $\mathbf z_i^\top$, respectively, and define the decay-weighted factors
\begin{equation}
    \begin{aligned}
        E_{i,:}
        &= (\boldsymbol{\gamma}_i\odot\mathbf u_i)^\top,\\
        \bar U_{i,:}
        &= (\mathbf u_i\oslash\boldsymbol{\gamma}_i)^\top,\\
        \bar R_{i,:}
        &= (\mathbf r_i\oslash\boldsymbol{\gamma}_i)^\top.
    \end{aligned}
    \label{eq:chunkwise_factors}
\end{equation}
where $\oslash$ denotes element-wise division. The causal interactions are
\begin{equation}
    \begin{aligned}
        L
        &=\operatorname{tril}
          \left(E(\bar U)^\top,-1\right),\\
        J
        &=\operatorname{tril}
          \left(E\bar R^\top,-1\right).
    \end{aligned}
    \label{eq:chunkwise_interactions}
\end{equation}
Here, the strictly lower triangular masks restrict interactions to earlier positions. Thus, $L\Delta$ and $JZ$ capture the effects of past delta corrections and residual writes, respectively, giving
\begin{equation}
    (I_C+L)\Delta
    =
    (I_C-J)Z-E\mathbf S_{\mathrm{in}}.
    \label{eq:chunkwise_system}
\end{equation}
The DPLR transition determines the coefficient matrix $I_C+L$, while the residual path adds $-JZ$ to the right-hand side.
Since $JZ$ depends only on inputs within the chunk, it can be precomputed independently of the incoming state and incorporated into the same triangular solve.

To separate the incoming-state dependence, we define the erase-side and write-side WY auxiliaries $\mathcal W$ and $\mathcal U$ through
\begin{equation}
    (I_C+L)
    \begin{bmatrix}
        \mathcal W & \mathcal U
    \end{bmatrix}
    =
    \begin{bmatrix}
        E & (I_C-J)Z
    \end{bmatrix}.
    \label{eq:chunkwise_auxiliaries}
\end{equation}
Both auxiliaries depend only on chunk inputs and can therefore be computed independently across chunks. Once the incoming state is available, the corrections are recovered as $\Delta=\mathcal U-\mathcal W\mathbf S_{\mathrm{in}}$.

The final state and all token outputs within the chunk are computed by aggregating contributions from the incoming state and both update paths. Define the factors propagated to the chunk boundary as $U_{\mathrm{tail}} =\bar U\operatorname{Diag}(\boldsymbol{\gamma}_C)$ and $R_{\mathrm{tail}} =\bar R\operatorname{Diag}(\boldsymbol{\gamma}_C)$.
For output computation, let $(Q_\gamma)_{i,:} =(\boldsymbol{\gamma}_i\odot\mathbf q_i)^\top$ and define
\begin{equation}
    \begin{aligned}
        P_u
        &=\operatorname{tril}
          \left(Q_\gamma(\bar U)^\top\right),\\
        P_r
        &=\operatorname{tril}
          \left(Q_\gamma\bar R^\top\right).
    \end{aligned}
    \label{eq:chunkwise_output_interactions}
\end{equation}
These masks include the diagonal because each output reads the state after the current update. The outgoing state and output block are then
\begin{equation}
    \begin{aligned}
        \mathbf S_{\mathrm{out}}
        &=
        \operatorname{Diag}(\boldsymbol{\gamma}_C)\mathbf S_{\mathrm{in}}
        +(U_{\mathrm{tail}})^\top\Delta
        +R_{\mathrm{tail}}^\top Z,\\
        O
        &=
        Q_\gamma\mathbf S_{\mathrm{in}}
        +P_u\Delta
        +P_rZ,
    \end{aligned}
    \label{eq:chunkwise_state_output}
\end{equation}
where row $i$ of $O$ is $\mathbf o_i^\top$. Together, these expressions complete the forward computation within each chunk, yielding all token outputs and the state passed to the next chunk.

Together, Eqs.~\eqref{eq:chunkwise_auxiliaries} and~\eqref{eq:chunkwise_state_output} organize the complete \model{} update into the standard chunkwise parallel framework. The auxiliary matrices can be prepared independently for each chunk, followed by recurrent state propagation across chunks and parallel computation of the output blocks. Within this schedule, residual writing contributes additional matrix products to the correction, state, and output computations while retaining the same $C\times C$ triangular system size. \model{} can therefore reuse the chunkwise training structure of delta-rule models, with linear complexity in sequence length for fixed chunk size and head dimensions.

\section{Experiments}

\subsection{Experimental Setup}
\label{sec:experimental_setup}

\subsubsection{Datasets}
In this work, we train models through 64K-context pretraining on Prolong dataset~\cite{prolong}.
We merge the 64K and 512K ProLong data, retokenize the combined corpus with the TinyLlama~\cite{zhang2024tinyllama} tokenizer, and repack it into uniform sequences of 64K tokens.
The resulting training stream contains approximately 82.78B tokens and uses next-token prediction over all non-padding tokens.
Recurrent states are reset at each training-sample boundary and propagate continuously across documents within each packed sequence.

\subsubsection{Baselines}

We compare \model{} with four baselines under the same training protocol.
KDA~\cite{kda} provides a direct reference for evaluating our extensions to the symmetric delta update, while GDN-2~\cite{GDN2} enables comparison with independent channel-wise gating under asymmetric erasure.
D-Clip~\cite{d_clip} caps each decay factor at $1/\|\mathbf A_t\|_2$, making the full transition non-expansive through stronger decay while preserving the original erase gate.
$\beta$-Guard~\cite{hatamizadeh2026betaguard} instead shrinks channel-wise deviations of the erase gate from its key-weighted mean when needed to satisfy a sufficient condition for non-expansive transitions, preserving both this mean and the original decay.
Both methods retain the original GDN-2 write term.
In comparison, \model{} combines symmetric erasure with decay-regulated residual writing, as described in Section~\ref{sec:stable_update}.

\subsubsection{Evaluation and Implementation}
\textbf{Evaluation.}
Following~\cite{GDN2}, we assess zero-shot common-sense reasoning on nine tasks: PIQA~\cite{bisk2020piqa}, HellaSwag~\cite{zellers-etal-2019-hellaswag}, WinoGrande~\cite{sakaguchi2020winogrande}, ARC-C and ARC-E~\cite{clark2018arc}, OpenBookQA~\cite{mihaylov-etal-2018-suit}, SocialIQA~\cite{sap-etal-2019-social}, BoolQ~\cite{clark-etal-2019-boolq}, and LAMBADA~\cite{paperno-etal-2016-lambada}.
We use normalized accuracy for PIQA, HellaSwag, ARC-C, ARC-E, and OpenBookQA, and accuracy for the remaining tasks.
For language modeling, we report WikiText~\cite{merity2016pointer} perplexity at 8K, 16K, 32K, and 64K windows, together with LAMBADA perplexity.
We evaluate long-context retrieval capabilities on 11 task types from RULER~\cite{hsieh2024ruler}. 
Specifically, S1, S2, S3, and MK1 correspond to \texttt{niah\_single\_1}, \texttt{niah\_single\_2}, \texttt{niah\_single\_3}, and \texttt{niah\_multikey\_1}, respectively. The other tasks include Multi-Query (MQ), Multi-Value (MV), Common-Word Extraction (CWE), Frequent-Word Extraction (FWE), Question Answering on HotpotQA (QA-H) and SQuAD (QA-S), and Variable Tracking (VT). 
Evaluation uses context lengths of 32K, 64K, and 128K.

\begin{table*}[t]
\centering
\caption{Performance comparison on zero-shot common-sense reasoning and language modeling. 
Bold and underlined values indicate the best and second-best results
within each setting, respectively.
}
\label{tab:short_tasks}
\setlength{\tabcolsep}{3pt}
\renewcommand{\arraystretch}{1.25}
\small

\resizebox{\textwidth}{!}{%
\begin{tabular}{@{}l >{\columncolor{black!3}}c *{14}{c}@{}}
\toprule
\multirow{2}{*}{Method}
& \multicolumn{10}{c}{ Accuracy (\%) $\uparrow$}
& \multicolumn{4}{c}{WikiText PPL $\downarrow$}
& \multirow{2}{*}{\makecell{LAMBADA\\PPL $\downarrow$}} \\

\cmidrule(lr){2-11}\cmidrule(lr){12-15}
& \textbf{Average}
& PIQA
& \makecell{Hella\\Swag}
& \makecell{Wino\\Grande}
& ARC-C
& ARC-E
& \makecell{Open\\BookQA}
& \makecell{Social\\IQA}
& BoolQ
& LAMBADA
& 8K & 16K & 32K & 64K & \\
\midrule

\multicolumn{16}{@{}l}{\textit{Recurrent Setting}} \\
\addlinespace[2pt]

\quad GDN-2
& \textbf{50.39}
& 71.27 & \underline{52.21} & 56.35
& \underline{30.29} & 54.42 & \underline{35.20}
& 40.12 & \underline{58.04} & \textbf{55.56}
& 7.849 & 7.838 & 7.825 & 7.804
& \textbf{7.856} \\


\quad D-Clip
& \underline{49.10}
& 70.67 & 51.65 & 55.96
& 28.41 & 53.11 & 34.60
& \textbf{41.76} & 51.99 & \underline{53.72}
& 7.889 & 7.877 & 7.873 & 7.871
& \underline{8.691} \\

\quad $\beta$-Guard
& 48.89
& 70.29 & 51.46 & \textbf{57.46}
& 29.69 & \underline{55.05} & 32.80
& \underline{41.15} & 50.89 & 51.25
& 7.858 & 7.835 & 7.820 & 7.806
& 9.543 \\

\quad KDA
& 49.02
& \textbf{71.60} & \textbf{52.95} & \underline{56.99}
& \textbf{31.31} & 53.70 & \textbf{37.00}
& 40.53 & 44.13 & 53.00
& \textbf{7.697} & \textbf{7.695} & \textbf{7.688} & \textbf{7.677}
& 9.250 \\

\midrule
\rowcolor{black!7}
\quad \textbf{Ours}
& \cellcolor{black!12}\textbf{50.39}
& \underline{71.44} & 51.57 & 54.93
& 30.20 & \textbf{58.71} & 34.80
& 39.87 & \textbf{59.69} & 52.28
& \underline{7.801} & \underline{7.788}
& \underline{7.776} & \underline{7.759}
& 9.333 \\

\midrule
\multicolumn{16}{@{}l}{\textit{Hybrid Setting}} \\
\addlinespace[2pt]

\quad GDN-2
& 49.91
& 70.95 & 52.05 & \textbf{56.91}
& 28.67 & 53.75 & \underline{34.80}
& \underline{41.56} & 56.91 & 53.64
& 7.546 & \textbf{7.507} & \textbf{7.495} & \textbf{7.493}
& 9.001 \\


\quad D-Clip
& 50.29
& \textbf{72.09} & \underline{52.50} & 55.96
& \textbf{31.31} & \textbf{56.27} & \textbf{35.00}
& 41.10 & 54.07 & 54.30
& \textbf{7.542} & 7.515 & 7.518 & 7.519
& \underline{8.277} \\

\quad $\beta$-Guard
& 49.68
& 71.00 & 51.76 & 54.46
& 30.29 & 52.31 & 32.60
& 41.45 & \underline{57.83} & 55.42
& 7.601 & 7.572 & 7.562 & 7.561
& \textbf{8.153} \\

\quad KDA
& \underline{50.36}
& \underline{71.65} & \textbf{52.59} & \underline{56.43}
& \underline{30.63} & 55.64 & 34.00
& 41.20 & 55.38 & \textbf{55.75}
& \underline{7.544} & \underline{7.510} & \underline{7.504} & \underline{7.504}
& 8.399 \\

\midrule
\rowcolor{black!7}
\quad \textbf{Ours}
& \cellcolor{black!12}\textbf{50.53}
& 70.73 & 52.29 & 54.06
& 29.78 & \underline{56.23} & \textbf{35.00}
& \textbf{41.66} & \textbf{59.42} & \underline{55.64}
& 7.594 & 7.564 & 7.557 & 7.558
& 8.586 \\

\bottomrule
\end{tabular}%
}
\end{table*}

\textbf{Implementation.}
We consider two distinct 20-layer architectures with different attention configurations. The first is a recurrent model consisting exclusively of linear-attention layers, while the second adopts a hybrid design that repeats three linear-attention layers followed by one full-attention layer five times.
Full attention covers the entire causal context.
Both architectures use a hidden size of 2,304 and an FFN intermediate size of 6,208.
Each linear-attention layer has 16 heads with 128-dimensional keys and values.
GDN-2 and \model{} have equal parameter counts within each architecture, approximately 1.60B for the recurrent model and 1.55B for the hybrid model.

All main-comparison models are randomly initialized and trained for one data epoch, comprising 39,473 optimizer steps with a global batch of 32 sequences, or approximately 2.10M token positions per step.
We use AdamW with $\beta_1=0.9$, $\beta_2=0.95$, weight decay 0.1, and gradient clipping at 1.0.
The learning rate increases linearly over 500 warmup steps and then remains constant at $3\times10^{-4}$.
Training uses BF16 mixed precision and full activation recomputation on 16 NVIDIA H800 GPUs per run.
Main performance comparisons use initialization seed 42 and data-order seed 2026, and evaluate the final checkpoint at step 39,473.

\subsection{Main Results}
To demonstrate the effectiveness of \model, we compare it against four representative baselines, including GDN-2, KDA, D-Clip, and $\beta$-Guard, across both Recurrent and Hybrid architectures. We evaluate their downstream task performance, language modeling quality, and long-context retrieval capabilities.
The results are reported in Table~\ref{tab:short_tasks} and~\ref{tab:ruler_wide}.

\begin{table*}[t]
\centering
\caption{RULER benchmark results (\%) at 32K, 64K, and 128K context lengths. NIAH denotes the average score across four needle-in-a-haystack tasks (S1, S2, S3, and MK1). The remaining metrics cover multi-query (MQ), multi-value (MV), common-word extraction (CWE), frequent-word extraction (FWE), question answering on HotpotQA (QA-H) and SQuAD (QA-S), and variable tracking (VT).
}
\label{tab:ruler_wide}
\vspace{0.5em}
\begingroup
\setlength{\tabcolsep}{3pt}
\renewcommand{\arraystretch}{1.06}
\fontsize{7.5}{8.8}\selectfont

\begin{tabularx}{\textwidth}{
  @{}l|c|
  >{\columncolor{black!4}\centering\arraybackslash}X|
  >{\centering\arraybackslash}X|
  *{7}{>{\centering\arraybackslash}X}@{}
}
\toprule
\multicolumn{11}{@{}l}{\textit{Recurrent Setting}} \\
\midrule
\addlinespace[2pt]
\textbf{Model} & \textbf{Seq Len} & \textbf{Avg.}
& \textbf{NIAH} & \textbf{MQ} & \textbf{MV}
& \textbf{CWE} & \textbf{FWE} & \textbf{QA-H}
& \textbf{QA-S} & \textbf{VT} \\
\midrule

\multirow{3}{*}{GDN-2}
& 32K  & 15.85 & 28.65 & 4.40 & 10.00 & 0.20 & 19.73 & \textbf{12.60} & \underline{7.95} & \textbf{4.84} \\
& 64K  & 12.95 & 24.45 & 2.25 & 4.30 & 0.20 & 13.33 & \textbf{11.60} & \textbf{6.95} & 6.04 \\
& 128K & 4.82 & 2.35 & \textbf{2.40} & \underline{2.55} & 0.32 & 12.87 & \underline{11.80} & \underline{6.15} & \underline{7.48} \\
\midrule

\multirow{3}{*}{D-Clip}
& 32K  & \underline{17.34} & 27.80 & \underline{5.15} & \underline{11.40} & 5.40 & \textbf{35.40} & \textbf{12.60} & 6.93 & 2.68 \\
& 64K  & 11.94 & 15.75 & \underline{3.35} & \textbf{5.45} & 1.60 & \textbf{34.13} & 10.60 & 6.15 & \textbf{7.08} \\
& 128K & 7.02 & 6.45 & 2.25 & \textbf{3.00} & \underline{0.50} & \textbf{29.53} & 10.60 & 5.42 & 0.12 \\
\midrule

\multirow{3}{*}{$\beta$-Guard}
& 32K  & 16.29 & \underline{29.60} & 4.05 & \textbf{11.90} & \underline{6.20} & 16.33 & \underline{11.60} & 7.40 & \underline{3.28} \\
& 64K  & \underline{15.27} & \underline{28.25} & 2.15 & 3.05 & \underline{2.04} & 25.80 & \underline{11.40} & 6.55 & 4.00 \\
& 128K & \underline{10.82} & \textbf{17.95} & 1.70 & 2.40 & 0.32 & 18.20 & \textbf{12.00} & 5.93 & 6.68 \\
\midrule

\multirow{3}{*}{KDA}
& 32K  & 16.51 & 27.75 & 2.55 & 5.70 & \textbf{9.62} & \underline{31.33} & 11.40 & 7.60 & 2.40 \\
& 64K  & 14.55 & 27.05 & 2.00 & 2.85 & \textbf{3.14} & 24.47 & 8.60 & \underline{6.87} & 3.88 \\
& 128K & 9.96 & \underline{17.85} & \underline{2.30} & 2.35 & \textbf{1.24} & 11.53 & 8.80 & 6.07 & 5.84 \\
\midrule

\rowcolor{black!7}
& 32K
& \cellcolor{black!12}\textbf{17.35}
& \textbf{29.90} & \textbf{7.15} & 8.80
& 2.70 & 30.13 & 10.40 & \textbf{8.80} & \underline{3.28} \\
\rowcolor{black!7}
& 64K
& \cellcolor{black!12}\textbf{16.34}
& \textbf{29.10} & \textbf{3.50} & \underline{5.10}
& 0.48 & \underline{32.33} & 9.60 & 6.13 & \underline{6.24} \\
\rowcolor{black!7}
\multirow{-3}{*}{\textbf{Ours}}
& 128K
& \cellcolor{black!12}\textbf{11.62}
& 17.30 & \textbf{2.40} & \underline{2.55}
& 0.40 & \underline{27.20} & 10.80 & \textbf{6.42} & \textbf{8.88} \\

\midrule
\multicolumn{11}{@{}l}{\textit{Hybrid Setting}} \\
\midrule
\addlinespace[2pt]
\textbf{Model} & \textbf{Seq Len} & \textbf{Avg.}
& \textbf{NIAH} & \textbf{MQ} & \textbf{MV}
& \textbf{CWE} & \textbf{FWE} & \textbf{QA-H}
& \textbf{QA-S} & \textbf{VT} \\
\midrule

\multirow{3}{*}{GDN-2}
& 32K  & 25.39 & 50.80 & 18.70 & \underline{23.20} & 0.36 & 4.53 & 19.00 & \textbf{10.27} & 0.00 \\
& 64K  & 15.50 & 27.20 & 10.15 & 17.70 & 0.08 & 14.33 & 12.40 & \underline{7.07} & 0.00 \\
& 128K & 8.57 & 11.85 & 2.00 & 6.50 & 0.02 & 20.27 & 11.80 & \underline{6.33} & 0.00 \\
\midrule

\multirow{3}{*}{D-Clip}
& 32K  & \underline{30.37} & \textbf{62.00} & \textbf{28.10} & \textbf{25.50} & \underline{2.94} & 4.60 & 17.20 & 6.73 & 1.00 \\
& 64K  & 18.65 & \textbf{33.30} & \textbf{18.25} & \textbf{20.20} & \underline{0.56} & 11.93 & 12.80 & 5.37 & \underline{2.84} \\
& 128K & \underline{8.85} & \textbf{13.70} & \textbf{10.40} & \textbf{12.10} & \underline{0.16} & 4.07 & 10.80 & 4.93 & \underline{0.08} \\
\midrule

\multirow{3}{*}{$\beta$-Guard}
& 32K  & 27.40 & 51.45 & 16.70 & 19.85 & 0.06 & \textbf{25.87} & \underline{21.00} & \underline{8.87} & 3.24 \\
& 64K  & \textbf{19.13} & \underline{30.90} & 12.95 & 18.60 & 0.04 & \textbf{32.33} & \underline{14.00} & 7.00 & 1.92 \\
& 128K & \textbf{10.52} & \underline{13.30} & 6.50 & 10.50 & 0.04 & \underline{25.27} & \underline{13.00} & \textbf{7.22} & 0.00 \\
\midrule

\multirow{3}{*}{KDA}
& 32K  & 24.31 & 42.55 & 16.85 & 22.35 & 1.92 & \underline{22.47} & \textbf{21.40} & 6.93 & \underline{5.28} \\
& 64K  & 14.49 & 21.95 & 9.80 & 13.45 & 0.30 & \underline{24.47} & \textbf{14.80} & \textbf{7.27} & 1.48 \\
& 128K & 8.20 & 8.00 & 2.65 & 4.20 & 0.08 & \textbf{31.73} & \textbf{14.20} & 5.22 & \underline{0.08} \\
\midrule

\rowcolor{black!7}
& 32K
& \cellcolor{black!12}\textbf{31.72}
& \underline{59.05} & \underline{22.60} & 22.55
& \textbf{4.34} & 12.67 & 12.20 & 6.47 & \textbf{31.88} \\
\rowcolor{black!7}
& 64K
& \cellcolor{black!12}\underline{18.69}
& 29.90 & \underline{17.10} & \underline{18.90}
& \textbf{1.30} & 14.53 & 10.80 & 5.57 & \textbf{17.80} \\
\rowcolor{black!7}
\multirow{-3}{*}{\textbf{Ours}}
& 128K
& \cellcolor{black!12}8.78
& 11.75 & \underline{8.10} & \underline{10.80}
& \textbf{0.72} & 13.53 & 10.00 & 3.67 & \textbf{2.80} \\

\bottomrule
\end{tabularx}
\endgroup
\end{table*}

\textbf{Downstream Performance and Language Modeling.} Table~\ref{tab:short_tasks} reports the downstream task accuracy and language modeling perplexity of different methods under the Recurrent and Hybrid settings.
Under the Recurrent setting, our method achieves an average downstream accuracy of 50.39\%, matching the original GDN-2 while outperforming D-Clip, $\beta$-Guard, and KDA. On WikiText, our method consistently achieves lower perplexity than GDN-2, D-Clip, and $\beta$-Guard across all evaluated context windows from 8K to 64K, with PPLs of 7.801, 7.788, 7.776, and 7.759, respectively. KDA, however, achieves lower perplexity, indicating that different recurrent update designs exhibit distinct trade-offs in language modeling performance.

Under the Hybrid setting, our method achieves the highest average downstream accuracy of 50.53\%, improving over GDN-2 by 0.62 percentage points and slightly outperforming KDA (50.36\%) and D-Clip (50.29\%). On WikiText, our method exhibits slightly higher perplexity than GDN-2, D-Clip, and KDA, while remaining comparable to $\beta$-Guard, with relatively small differences across context windows.

\textbf{Long-Context Retrieval}. Table~\ref{tab:ruler_wide} evaluates the ability of different methods to retrieve and utilize historical information at context lengths of 32K, 64K, and 128K.
In the Recurrent setting, our method achieves the highest average RULER scores across all three context lengths, reaching 17.35, 16.34, and 11.62, respectively. Notably, at 128K, our method achieves 11.62, outperforming GDN-2 (4.82), D-Clip (7.02), $\beta$-Guard (10.82), and KDA (9.96). A similar trend is observed on NIAH, where our method achieves the best results at 32K and 64K, with scores of 29.90 and 29.10, respectively. At 128K, however, $\beta$-Guard and KDA slightly outperform our method.

For the Hybrid setting, our method achieves the highest average RULER score at 32K, reaching 31.72 and outperforming D-Clip (30.37) and the other baselines. At 64K, our method achieves 18.69, slightly below $\beta$-Guard (19.13), while outperforming D-Clip, GDN-2, and KDA. At 128K, $\beta$-Guard and D-Clip achieve slightly higher scores of 10.52 and 8.85, respectively, compared with 8.78 for our method. Furthermore, our method exhibits a particularly pronounced advantage on Variable Tracking, achieving scores of 31.88 and 17.80 at 32K and 64K, respectively, substantially outperforming all other baselines.
These results demonstrate that \model~maintains strong historical information retrieval capabilities across different context lengths and architectural settings, with particularly competitive long-context retrieval performance under the Recurrent setting.

\textbf{Overall Analysis}. Taken together, \model~achieves a favorable balance between recurrent state stability and modeling capability. Compared with stabilization methods such as D-Clip and $\beta$-Guard, which constrain state propagation or erasure strength, our method retains channel-wise gating and flexible information writing while achieving competitive performance across multiple language modeling and long-context retrieval benchmarks. In particular, under the Recurrent setting, our method consistently achieves the highest average RULER scores across all three evaluated context lengths.
These results suggest that effective control of recurrent state amplification can be achieved without sacrificing the model's ability to utilize long-term historical information through a carefully designed combination of state erasure and information writing mechanisms.

\begin{figure}[]
    \centering
\includegraphics[width=0.99\linewidth]{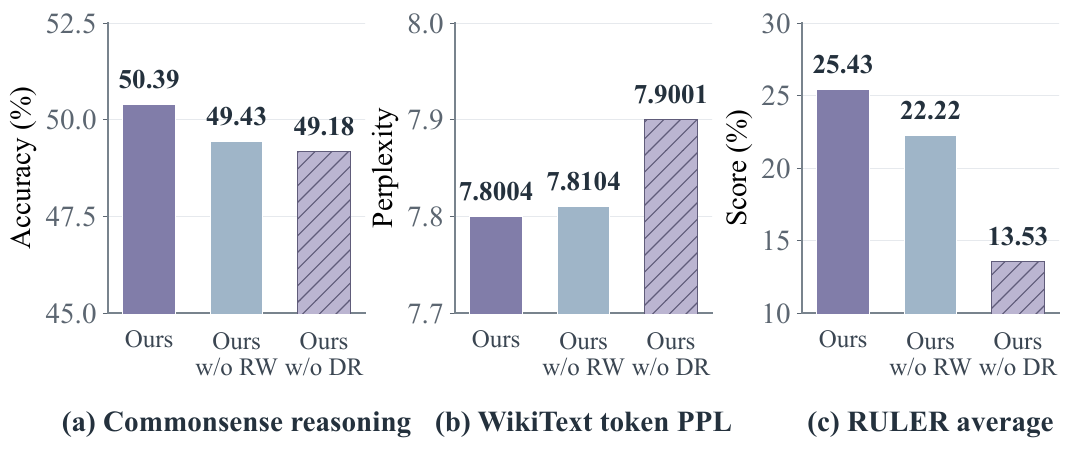}
    \vspace{-1em}
    \caption{Results for ablation studies.}
    \label{fig:ablation}
    \vspace{-1.5em}
\end{figure}

\textbf{Ablation Studies.} 
To investigate the effectiveness of different components of \model, we conduct ablation studies on different components.
Figure~\ref{fig:ablation} presents ablation results to examine the contributions of residual writing (RW) and decay regulation (DR).
\emph{Ours w/o RW} removes residual writing by setting $\mathbf r_t=\mathbf0$ in Eq.~\eqref{eq:stable_GDN-2}, yielding $\mathbf S_t=\bigl[\mathbf I-\mathbf u_t(\mathbf u_t)^\top\bigr]\mathbf D_t\mathbf S_{t-1}+\mathbf u_t\mathbf z_t^\top$.
\emph{Ours w/o DR} removes the factor $\sqrt{1-\mathbf d_t^2}$ from the residual write, giving $\mathbf r_t=\mathbf k_t-\mathbf u_t$. The total write key then becomes $\mathbf k_t$, resulting in $\mathbf S_t=\bigl[\mathbf I-\mathbf u_t(\mathbf u_t)^\top\bigr]\mathbf D_t\mathbf S_{t-1}+\mathbf k_t\mathbf z_t^\top$.
Figure~\ref{fig:ablation}(a) reports the average zero-shot accuracy across the nine benchmarks in Table~\ref{tab:short_tasks}, while Figure~\ref{fig:ablation}(b) reports WikiText-2 token perplexity, both with an 8K context. Figure~\ref{fig:ablation}(c) reports the mean score over four RULER retrieval tasks (\texttt{NIAH single-1/2/3} and \texttt{multikey-1}) at 32K, 64K, and 128K context lengths, averaging all 12 task–length combinations equally.

Removing either component consistently degrades performance across the evaluated metrics. Specifically, removing RW reduces the short-task average accuracy from 50.39\% to 49.43\% and the RULER score from 25.43 to 22.22, accompanied by an increase in WikiText perplexity from 7.8004 to 7.8104. This demonstrates the importance of residual writing in recovering writing capacity suppressed by symmetric erasure. Removing DR leads to a more pronounced degradation in long-context retrieval, with the RULER score dropping to 13.53, while also reducing short-task accuracy to 49.18\%. These results highlight the importance of regulating residual writes to control their cumulative effects and preserve long-term memory capabilities. 
Overall, the two components play complementary roles, with residual writing restoring flexible memory updates and decay regulation controlling the additional energy introduced by this pathway.

\section{Related Work}

\subsection{Linear Attention and Delta-rule Models}

Recent advances in linear attention have improved sequence modeling through more expressive gating mechanisms and recurrent state updates~\cite{gla,GDN,kda,sun2026fg2gdn,GDN2}.
The basic formulation uses kernel feature maps to summarize the prefix in a recurrent state of key--value outer products, allowing queries to retrieve information at a cost independent of context length~\cite{katharopoulos2020transformers}.
Interpreting this recurrence as a fast-weight programmer~\cite{schlag2021linear} motivates replacing additive writes with delta-rule updates that selectively revise existing associations~\cite{irie2021going}, while chunkwise parallel algorithms make these updates practical for training at scale~\cite{yang2024parallelizing}. A parallel thread governs how long the state retains its contents, through fixed exponential decay in RetNet~\cite{retnet}, input-dependent per-channel decay in gated linear attention~\cite{gla}, depth-wise gate floors and state expansion in HGRN and HGRN2~\cite{HGRN,qin2024hgrn2}, and selective discretizations in the Mamba line~\cite{mamba,ssm,mamba3}. Gated DeltaNet applied the delta correction to a decayed state~\cite{GDN}, and Kimi Delta Attention refined the decay to a diagonal~\cite{kda}. 
GDN-2~\cite{GDN2} then split the shared scalar gate into independent channel-wise erase and write gates, an asymmetry EDA approaches from the write-address side~\cite{li2026erasethendeltaattentiondecouplingerase}, while DeltaProduct~\cite{DeltaProduct} and RWKV-7~\cite{rwkv7} pursue expressive transitions by other means. 
In our work, we identify recurrent state amplification as a fundamental source of instability in GDN-2 and propose \model{}, which extends the symmetric delta update of KDA while preserving non-expansive recurrent state propagation.

\subsection{Memory Capacity and Stabilization of Recurrent Models
}

A fixed-size state bounds how much history a recurrent model can carry. Attention reads as associative memory~\cite{NEURIPS2023_0561738a}; capacity rather than parameter count separates attention from gated convolution, and the resulting recall–throughput trade-off can be quantified~\cite{Zoology,pmlr-v235-arora24a}; no fixed latent state reproduces arbitrary inputs verbatim~\cite{pmlr-v235-jelassi24a}. Designs respond by widening the state~\cite{qin2024hgrn2}, letting it grow logarithmically~\cite{guo2026loglinearattention}, or externalizing it~\cite{NEURIPS2025_a4ca07aa}. Whether a state of a given size propagates without numerical degradation is a separate question with older roots: long-range learning is tied to products of per-step Jacobians~\cite{Bengio_1994}, gradients are rescaled by their norm~\cite{pmlr-v28-pascanu13}, and transitions may be confined to unitary matrices~\cite{pmlr-v48-arjovsky16}. In transformers the analogue targets logits rather than states, via query–key normalization~\cite{ICLR2024_d848cb2c} or direct bounds on how far logits may move in one step~\cite{anson2025controllingchangesattentionlogits}. Each of these constrains a single transition; accumulation is left implicit, which suffices for normal transitions but not for the non-symmetric erase matrix of GDN-2. 
In our work, we show that non-expansive recurrent state propagation in \model{} can be guaranteed through symmetric erasure, preventing cumulative amplification while preserving flexible memory updates.

\section{Conclusion}
In this work, we identify asymmetric erasure as a fundamental source of recurrent state amplification in GDN-2, showing theoretically how non-normal state transitions can accumulate across tokens and lead to numerical instability. To address this issue, we propose \model{}, which extends the symmetric delta update of KDA with channel-wise erase and write gates and decay-regulated residual writing, ensuring non-expansive state propagation while preserving flexible memory updates and chunkwise parallel computation.
Extensive experiments demonstrate that our method stabilizes long-context training and delivers strong long-context retrieval and downstream task performance, while maintaining competitive language modeling quality.
These findings highlight the importance of controlling recurrent state dynamics and demonstrate that numerical stability can be substantially improved without sacrificing the expressive capacity of recurrent memory.

\bibliography{reference}
\bibliographystyle{icml2025}

\newpage
\appendix
\onecolumn

\appendix

\twocolumn[{
    \renewcommand\twocolumn[1][]{#1}
    \begin{center}
      \textbf{\fontsize{15}{48}\selectfont Appendix}
    \end{center}
    \vspace{0.5cm}
}]

\section{Proofs for the Theoretical Analysis}
\label{app:GDN-2_theory}

\subsection{Single-Step Erasure Gain}
\label{app:GDN-2_single_step}

\begin{proof}[Proof of Theorem~\ref{thm:GDN-2_single_step}]
Under the assumptions of the theorem,
Eq.~\eqref{eq:gated_key_decomposition} gives
\begin{equation}
    \mu=\sum_{i=1}^{d_k}b_i k_i^2\in[0,1].
    \label{eq:app_GDN-2_mu_range}
\end{equation}
The second column of $\mathbf{T}$ in
Eq.~\eqref{eq:GDN-2_erase_geometry} has norm
$\sqrt{1+\gamma^2}\geq1$.
Together with the identity action on the orthogonal complement,
this implies $\|\mathbf{A}\|_2=\|\mathbf{T}\|_2$.

Let $\sigma_1\geq\sigma_2\geq0$ be the singular values of
$\mathbf{T}$. Their squared sum and product satisfy
\begin{equation}
    \begin{aligned}
        \sigma_1^2+\sigma_2^2
        &=\|\mathbf{T}\|_F^2=(1-\mu)^2+\gamma^2+1,\\
        \sigma_1\sigma_2
        &=|\det\mathbf{T}|=1-\mu.
    \end{aligned}
    \label{eq:app_GDN-2_singular_identities}
\end{equation}
Consequently,
\begin{equation}
    \begin{aligned}
        (\sigma_1+\sigma_2)^2&=(2-\mu)^2+\gamma^2,\\
        (\sigma_1-\sigma_2)^2&=\mu^2+\gamma^2.
    \end{aligned}
\end{equation}
Taking the nonnegative square roots yields
\begin{equation}
    \|\mathbf{A}\|_2=\sigma_1
    =\frac{\sqrt{\mu^2+\gamma^2}
    +\sqrt{(2-\mu)^2+\gamma^2}}{2},
\end{equation}
which proves Eq.~\eqref{eq:GDN-2_erasure_norm}.
When $\gamma=0$, this expression equals $1$ because
$\mu\in[0,1]$. When $\gamma>0$, the two square roots strictly
exceed $\mu$ and $2-\mu$, respectively, giving
$\|\mathbf{A}\|_2>1$. Hence $\|\mathbf{A}\|_2=1$ if and only if
$\gamma=0$, which is equivalent to $\mathbf{a}=\mu\mathbf{k}$
and therefore to collinearity of $\mathbf{a}$ and $\mathbf{k}$.
\end{proof}

The same representation also verifies that $\mathbf{A}$ is non-normal
when $\gamma>0$, since
\begin{equation}
    \mathbf{T}^{\top}\mathbf{T}-\mathbf{T}\mathbf{T}^{\top}
    =\begin{pmatrix}
        -\gamma^2 & \mu\gamma \\
        \mu\gamma & \gamma^2
    \end{pmatrix}\neq\mathbf{0}
    \quad\text{when }\gamma>0.
    \label{eq:app_GDN-2_non_normality}
\end{equation}
For the recurrent state energy introduced in
Section~\ref{sec:theoretical_analysis}, the maximum amplification factor is
\begin{equation}
    \sup_{\mathbf{S}\neq\mathbf{0}}
    \frac{E(\mathbf{A}\mathbf{S})}{E(\mathbf{S})}
    =\|\mathbf{A}\|_2^2.
    \label{eq:app_GDN-2_energy_gain}
\end{equation}
The upper bound follows by applying the operator norm bound to
each column of $\mathbf{S}$. Equality is attained when one column
lies along a leading right singular direction of $\mathbf{A}$
and all other columns are zero. Thus, nonzero shear permits
erasure alone to increase the energy of the recurrent state,
without any new write.

\subsection{Unbounded Growth Without Further Writes}
\label{app:GDN-2_growth_construction}

\begin{proof}[Proof of Corollary~\ref{cor:GDN-2_state_growth}]
A periodic sequence of two alternating transitions realizes the
alignment described in Section~\ref{sec:theoretical_analysis}.
We construct it on two key coordinates and one recurrent state column,
then embed it in higher dimensions.
For this example, choose $\mathbf{D}_t=d\mathbf{I}$ with $d=0.99$
at every token.

Choose the following pairs of keys and erase gates,
\begin{equation}
    \begin{aligned}
        \mathbf{k}_a&=\frac{1}{\sqrt{5}}\begin{pmatrix}2\\1\end{pmatrix},
        &\mathbf{b}_a&=\frac{1}{5}\begin{pmatrix}1\\4\end{pmatrix},\\
        \mathbf{k}_b&=\frac{1}{\sqrt{5}}\begin{pmatrix}1\\2\end{pmatrix},
        &\mathbf{b}_b&=\frac{1}{5}\begin{pmatrix}4\\1\end{pmatrix}.
    \end{aligned}
    \label{eq:app_GDN-2_periodic_parameters}
\end{equation}
Both keys have unit norm, and all erase gate entries lie in $(0,1)$.
The corresponding erase operators are
\begin{equation}
    \begin{aligned}
        \mathbf{A}_a&=\frac{1}{25}
        \begin{pmatrix}21&-8\\-2&21\end{pmatrix},\\
        \mathbf{A}_b&=\frac{1}{25}
        \begin{pmatrix}21&-2\\-8&21\end{pmatrix}
        =\mathbf{A}_a^{\top}.
    \end{aligned}
    \label{eq:app_GDN-2_transpose_erasure}
\end{equation}
Hence the full transitions
$\mathbf{M}_a=d\mathbf{A}_a$ and
$\mathbf{M}_b=d\mathbf{A}_b$ satisfy
$\mathbf{M}_b=\mathbf{M}_a^{\top}$.
For $\mathbf{A}_a$, the geometric coefficients are
$\mu=8/25$ and $\gamma=6/25$.
Theorem~\ref{thm:GDN-2_single_step} therefore gives
\begin{equation}
    \lambda=d^2\frac{19+6\sqrt{2}}{25}
    \approx1.077533>1.
    \label{eq:app_GDN-2_pair_gain}
\end{equation}

At the first token, choose the write
\begin{equation}
    \mathbf{S}_1=
    \frac{1}{\sqrt{100+10\sqrt{2}}}
    \begin{pmatrix}7\\-(1+5\sqrt{2})\end{pmatrix},
    \qquad\|\mathbf{S}_1\|_2=1.
    \label{eq:app_GDN-2_initial_write}
\end{equation}
This write is realized by
$\mathbf{k}_1=\mathbf{S}_1$,
$\mathbf{b}_1=(1/2,1/2)^{\top}$,
$w_1=1/2$, and $v_1=2$.
Indeed, the initial recurrent state is zero, so the first update
reduces to $\mathbf{k}_1(w_1v_1)=\mathbf{S}_1$.
For every subsequent token $t\geq2$, set $w_t=1/2$ and $v_t=0$,
so that no further content is written while the write gate remains
strictly between zero and one.
Alternate the pairs in
Eq.~\eqref{eq:app_GDN-2_periodic_parameters} so that
\begin{equation}
    \mathbf{M}_{2n+2}=\mathbf{M}_a,
    \qquad
    \mathbf{M}_{2n+3}=\mathbf{M}_b,
    \qquad n\geq0.
\end{equation}

Direct multiplication gives
\begin{equation}
    \mathbf{M}_b\mathbf{M}_a
    =\frac{d^2}{125}
    \begin{pmatrix}89&-42\\-42&101\end{pmatrix},
    \label{eq:app_GDN-2_pair_product}
\end{equation}
and the initial column in
Eq.~\eqref{eq:app_GDN-2_initial_write} satisfies
$\mathbf{M}_b\mathbf{M}_a\mathbf{S}_1=\lambda\mathbf{S}_1$.
It follows by induction that
\begin{equation}
    \begin{aligned}
        \mathbf{S}_{2n+1}&=\lambda^n\mathbf{S}_1,\\
        \mathbf{S}_{2n+2}&=\lambda^n\mathbf{M}_a\mathbf{S}_1,
        \qquad n\geq0.
    \end{aligned}
    \label{eq:app_GDN-2_periodic_trajectory}
\end{equation}
Moreover, the transpose pairing implies
\begin{equation}
    \|\mathbf{M}_a\mathbf{S}_1\|_2^2
    =\mathbf{S}_1^{\top}\mathbf{M}_b\mathbf{M}_a\mathbf{S}_1
    =\lambda.
\end{equation}
Since $\lambda=\sigma_{\max}(\mathbf{M}_a)^2$, this equality shows that
$\mathbf{S}_1$ lies along a leading right singular direction of
$\mathbf{M}_a$. The first transition maps it to the corresponding
left singular direction, and the second maps it back.
Each pair both amplifies the recurrent state and restores the
direction required for the next pair. In particular, for every $t\geq1$,
\begin{equation}
    \|\mathbf{S}_t\|_2=\lambda^{(t-1)/2},
    \qquad
    E(\mathbf{S}_t)=\lambda^{t-1}\longrightarrow\infty.
    \label{eq:app_GDN-2_all_step_growth}
\end{equation}
This growth occurs entirely after writing ceases, despite the
strict decay applied at every step.

For a recurrent state with more coordinates or columns, embed every
key, including $\mathbf{k}_1$, in the same two key coordinates and
set all remaining key entries to zero. Set the remaining
erase gate entries and all write gate entries to $1/2$, retain
$\mathbf{D}_t=d\mathbf{I}$, and place the initial value $2$ in
one value coordinate, with all other values zero. All subsequent value vectors are zero. The resulting recurrent state remains supported on the chosen two coordinates and one column, so
\begin{equation}
    \|\mathbf{S}_t\|_F=\lambda^{(t-1)/2}
    \longrightarrow\infty.
    \label{eq:app_GDN-2_embedded_growth}
\end{equation}
All keys remain normalized and all erase, write, and decay gate entries
lie in $(0,1)$, so the embedded sequence satisfies the corollary.
\end{proof}

\section{Details of Empirical Characterization}
\label{appendix:empirical}

\subsection{Training Protocol and Failure Accounting}
\label{app:empirical_protocol}

The Pure Linear and Hybrid architectures, tokenizer, pretraining data, and shared optimization settings follow Section~\ref{sec:experimental_setup}.
Pure Linear uses GDN-2 in all 20 layers, while Hybrid repeats three GDN-2 layers followed by one full attention layer.
Their parameter counts are approximately 1.595B and 1.554B, respectively.
Pretraining starts from random initialization and uses the ProLong corpus~\citep{prolong} with next token supervision over all nonpadding tokens.
The maximum packed sequence length is 65,536 tokens.
The recurrent state propagates across documents within each pretraining pack and is reset at the start of the next pack.

Post-training uses the official training split of \texttt{nvidia/ChatQA2-Long-SFT-data}~\citep{xu2025chatqa}, with loss applied only to answer and EOS targets.
The standalone post-training condition starts from random initialization, whereas post-training after pretraining starts from the final weights of the corresponding pretraining run.
The latter resets the optimizer, learning rate schedule, training random number state, and step counter.
We retain 143,367 complete QA examples that fit within 65,536 tokens, yielding approximately 2.170B nonpadding input tokens and 79.987M supervised targets.
Examples exceeding the length limit are excluded without truncation.
Each QA example is an independent segment within a pack, with positions, recurrent state, and convolution state reset at its boundary; full attention in Hybrid is also restricted to that segment.
Thus, the 64K pack limit does not imply uninterrupted recurrent state propagation across separate QA examples.
Both post-training conditions use one data epoch, a global batch of 16 packs, and 2,331 optimizer updates.
The learning rate warms up to $3\times10^{-4}$ over 100 updates, remains constant, and decays linearly to $3\times10^{-5}$ over the final 233 updates.
Other optimizer settings follow Section~\ref{sec:experimental_setup}.

The four seed configurations form a $2\times2$ design crossing data order seeds $\{2026,2027\}$ with initialization and training seeds $\{42,43\}$.
Seeds~1--4 correspond to the pairs $(2026,42)$, $(2026,43)$, $(2027,42)$, and $(2027,43)$, respectively.
The packing seed is fixed at 3407, so changing the data seed changes pack order rather than the underlying packing.
The same seed configuration identifies corresponding stages within each architecture.
A failed pretraining run does not launch its dependent post-training stage, leaving three evaluated stages for Pure Linear and two for Hybrid in the final column of Table~\ref{tab:GDN-2_training_failures}.

We count failures according to the loss NaN criterion in Section~\ref{sec:empirical_characterization}.
Table~\ref{tab:GDN-2_training_failures} uses one designated trajectory per seed configuration and training stage.
For Hybrid Seed~2, the initial standalone post-training trajectory completed all 2,331 updates, while a separately initialized repeat with the same seeds failed at update 471.
The table uses the latter as its designated trajectory, giving 3/4 failures for Hybrid post-training; the earlier completion is disclosed here as an additional trajectory.
Other repeats and failure replays do not contribute additional seed configurations, and resumption after an infrastructure interruption continues the same trajectory.
The designated runs comprise 14 completed stages and 7 numerical failures, with 3 dependent post-training stages not launched.

\subsection{Replay and Diagnostic Protocol}
\label{app:empirical_replay}

Figure~\ref{fig:GDN-2_training_diagnostics} examines the Pure Linear Seed~4 pretraining failure at update 2,058.
Replay starts from the last finite checkpoint, saved after update 2,057, and uses the triggering input sequence of 65,536 tokens, identified as sample 1,025,853 in the packed dataset.
The recurrent state diagnostics select layer~4 and head~5, with both indices starting from zero.
GDN and KDA are replayed using checkpoints at the same training step, the same input sample, and the corresponding layer and head.
Each model supplies its own keys, gates, and decay factors.

The ordinary replay records recurrent state norms and the occurrence of nonfinite values.
State observations are sampled at intervals of 64 tokens, with the first nonfinite GDN-2 state observed at token 52,736.
Nonfinite values then propagate through subsequent matrix computations to the logits and loss.
The overflow marker denotes this observed position, rather than a threshold applied to the Frobenius norm.
Training uses BF16 mixed precision, with recurrent state accumulation primarily in FP32 and states saved across chunks in BF16.
The precision of the accumulation and saved states is therefore distinguished in the diagnosis.

For Figure~\ref{fig:empirical_no_write}, each model first processes 64 tokens with ordinary writes to obtain its initial recurrent state.
From token 65 onward, the additive write term is set to zero while the keys, gates, and decay factors remain fixed to the values extracted from the corresponding ordinary replay.
This isolates the propagation in Eq.~\eqref{eq:GDN-2_state_expansion} without changing the transition sequence.
The offline diagnostic uses FP64 arithmetic with recurrent state rescaling and a separately accumulated logarithmic scale to track norms over the full sequence.
The extended logarithmic curves describe this propagation diagnostic; overflow timing is taken from the native replay in Figure~\ref{fig:empirical_state_norm}.
Effective rank in Figure~\ref{fig:empirical_rank} is evaluated on finite states from ordinary replay, with scalar normalization before singular value computation.
The ratio defined in the main text is invariant to this normalization and is also known as stable rank.

\begin{figure}[t]
    \centering
    \includegraphics[width=0.48\linewidth]{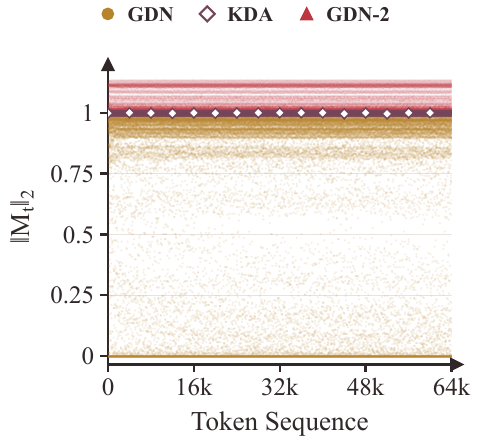}%
    \hfill
    \includegraphics[width=0.48\linewidth]{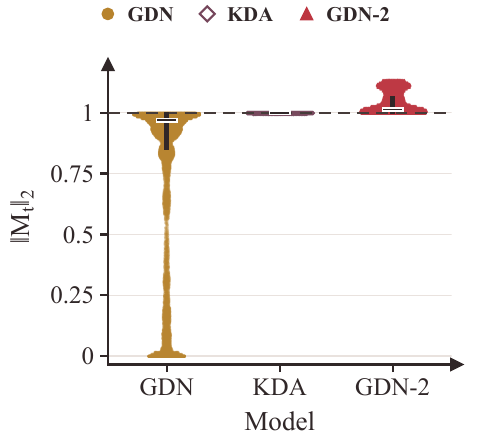}
    \vspace{-1em}
    \caption{
        Spectral norms of recurrent state transitions.
        The full transition includes erasure and decay.
        Measurements use the same input sequence and selected layer and head for GDN, KDA, and GDN-2, with 65,536 transitions per model.
        The left panel shows norms by token position, and the right panel shows their distributions with medians and interquartile ranges.
        The dashed line marks $\|\mathbf{M}_t\|_2=1$.
    }
    \label{fig:nonwriting_transition_norm}
    \vspace{-2em}
\end{figure}

Figure~\ref{fig:nonwriting_transition_norm} reports the spectral norms of the full transitions $\mathbf{M}_t$, including both erasure and decay.
For GDN-2 in this sample and selected head, the maximum is 1.135 and 89.58\% of the transitions have norm greater than one.
These measurements identify expanding directions in individual transitions, while the trajectory in Figure~\ref{fig:empirical_no_write} shows the accumulated amplification realized by their action on the recurrent state.

\subsection{Recurrent State Magnitudes Across Seeds}
\label{app:state_magnitudes}

Table~\ref{tab:GDN-2_seed_state_maxima} reports the largest recorded finite absolute recurrent state entry during each completed training trajectory.
These maxima are taken over available monitoring records; the detailed failure replay in Appendix~\ref{app:empirical_replay} is reported separately.
The seed mapping and designated trajectories follow Appendix~\ref{app:empirical_protocol}.

\begin{table}[t]
    \centering
    \small
    \caption{
        Maximum recurrent state magnitudes across seeds.
        Entries give the largest recorded finite absolute recurrent state value during each completed training trajectory, rounded to four significant digits.
        \xmark{} marks a run terminated by loss NaN.
        An em dash denotes a stage that was not launched.
    }
    \vspace{0.5em}
    \label{tab:GDN-2_seed_state_maxima}
    \footnotesize
    \setlength{\tabcolsep}{3pt}
    \begin{tabular}{@{}l|c|ccc@{}}
        \toprule
        Model & Seed & Pretraining & Post-training
        & \makecell[c]{Pretraining $\rightarrow$\\Post-training} \\
        \midrule
        \multirow{4}{*}{\makecell[l]{Pure\\Linear}}
        & 1 & $4.383\times10^{5}$  & $1.016\times10^{6}$  & $7.559\times10^{11}$ \\
        & 2 & $3.157\times10^{36}$ & \xmark               & $5.504\times10^{3}$  \\
        & 3 & $1.660\times10^{20}$ & $8.048\times10^{14}$ & $1.464\times10^{5}$  \\
        & 4 & \xmark               & $1.047\times10^{11}$ & \textemdash          \\
        \midrule
        \multirow{4}{*}{Hybrid}
        & 1 & $9.408\times10^{3}$  & \xmark               & $4.940\times10^{2}$  \\
        & 2 & \xmark               & \xmark               & \textemdash          \\
        & 3 & \xmark               & $3.707\times10^{5}$  & \textemdash          \\
        & 4 & $3.264\times10^{11}$ & \xmark               & $1.590\times10^{2}$  \\
        \bottomrule
    \end{tabular}
    \vspace{-1em}
\end{table}

Extreme values occur even along training trajectories that ultimately complete.
For example, Pure Linear Seed~2 completes pretraining after reaching a recorded magnitude of approximately $3.157\times10^{36}$ during training.
This value remains within the finite representable range of FP32, whose maximum is approximately $3.403\times10^{38}$.
The absence of loss NaN therefore does not imply controlled recurrent state magnitudes throughout training.

\section{State Bounds for \model}
\label{app:state_bounds}

We use the update, normalization, and gate parameterization specified in Sections~\ref{sec:stable_update}--\ref{sec:stability_guarantees} throughout.

\subsection{Write Norm and Additive Accumulation}
\label{app:write_norm}

The residual in Eq.~\eqref{eq:residual_write} gives, for each coordinate,
\begin{equation}
    (\mathbf u_t+\mathbf r_t)_i
    =\left[\sqrt{b_{t,i}}+
    \sqrt{1-d_{t,i}^2}(1-\sqrt{b_{t,i}})\right]k_{t,i}.
    \label{eq:app_write_coordinate}
\end{equation}
The coefficient in brackets lies in $[0,1]$, because both
$\sqrt{b_{t,i}}$ and $\sqrt{1-d_{t,i}^2}$ lie in that interval.
Hence $\|\mathbf u_t+\mathbf r_t\|_2\leq\|\mathbf k_t\|_2\leq1$, and the Frobenius norm of an outer product yields
\begin{equation}
    \begin{aligned}
        \|(\mathbf u_t+\mathbf r_t)\mathbf z_t^\top\|_F
        &=\|\mathbf u_t+\mathbf r_t\|_2\|\mathbf z_t\|_2\\
        &\leq\|\mathbf k_t\|_2\|\mathbf z_t\|_2
        =\|\mathbf k_t\mathbf z_t^\top\|_F.
    \end{aligned}
    \label{eq:app_write_norm}
\end{equation}
Expanding Eq.~\eqref{eq:stable_GDN-2}, applying the triangle inequality, and using
$\|\boldsymbol\Phi_{T:t+1}\|_2\leq1$
gives Eq.~\eqref{eq:stable_additive_bound}.
This argument accounts for arbitrary alignment between successive writes.

\subsection{Proof of the Recurrent State Norm Bound}
\label{app:input_energy}

\begin{proof}[Proof of Proposition~\ref{prop:write_energy_bound}]
Fix a token and omit its index.
Introduce the auxiliary quantities
\begin{equation}
    \mathbf f=\mathbf k-\mathbf u,
    \qquad
    \mathbf C=(\mathbf I-\mathbf D^2)^{1/2},
    \qquad
    \nu=\|\mathbf u\|_2^2.
\end{equation}
Equation~\eqref{eq:residual_write} gives $\mathbf r=\mathbf C\mathbf f$.
Since $\mathbf A^{\mathrm{sym}}$ is symmetric and
$\mathbf I-(\mathbf A^{\mathrm{sym}})^2=(2-\nu)\mathbf u(\mathbf u)^\top$,
\begin{equation}
    \begin{aligned}
        \mathbf I-\mathbf M\mathbf M^\top
        &=\mathbf I-(\mathbf A^{\mathrm{sym}})^2
        +\mathbf A^{\mathrm{sym}}(\mathbf I-\mathbf D^2)\mathbf A^{\mathrm{sym}}\\
        &=(2-\nu)\mathbf u(\mathbf u)^\top
        +\mathbf A^{\mathrm{sym}}\mathbf C^2\mathbf A^{\mathrm{sym}}
        =\mathbf L\mathbf L^\top,
    \end{aligned}
    \label{eq:app_energy_factorization}
\end{equation}
where the auxiliary matrix is
$\mathbf L=[\sqrt{2-\nu}\,\mathbf u\;\;\mathbf A^{\mathrm{sym}}\mathbf C]$.
The write vector admits a factorization through the same matrix,
\begin{equation}
    \mathbf u+\mathbf r=\mathbf L\boldsymbol\xi,
    \qquad
    \boldsymbol\xi=
    \begin{bmatrix}
        \displaystyle\frac{1+(\mathbf u)^\top\mathbf C\mathbf f}
        {\sqrt{2-\nu}}\\[4pt]
        \mathbf f
    \end{bmatrix}.
    \label{eq:app_write_factorization}
\end{equation}
Indeed, the first block contributes
$\mathbf u[1+(\mathbf u)^\top\mathbf C\mathbf f]$,
while the second contributes
$\mathbf A^{\mathrm{sym}}\mathbf C\mathbf f
=\mathbf C\mathbf f-\mathbf u(\mathbf u)^\top\mathbf C\mathbf f$.
Their sum is $\mathbf u+\mathbf r$.

To bound $\|\boldsymbol\xi\|_2$, set
$\eta=(\mathbf u)^\top\mathbf f$ and $\tau=\|\mathbf f\|_2^2$.
The coordinate products
$u_if_i=\sqrt{b_i}(1-\sqrt{b_i})k_i^2$ are nonnegative.
Since the diagonal entries of $\mathbf C$ lie in $[0,1]$,
\vspace{-0.1in}
\begin{equation}
    \begin{gathered}
        0\leq(\mathbf u)^\top\mathbf C\mathbf f\leq\eta,
        \qquad \eta^2\leq\nu\tau,\\
        \nu+2\eta+\tau=\|\mathbf k\|_2^2\leq1.
    \end{gathered}
\end{equation}
The denominator in Eq.~\eqref{eq:app_write_factorization} is positive because $\nu\leq1$.
Moreover,
\begin{equation}
    \begin{aligned}
        &\left(\frac32-\tau\right)(2-\nu)-(1+\eta)^2\\
        &\quad=2(1-\nu-2\eta-\tau)
        +\frac\nu2+2\eta\\
        &\qquad+(\nu\tau-\eta^2)\geq0.
    \end{aligned}
\end{equation}
It follows that
\begin{equation}
    \|\boldsymbol\xi\|_2^2
    \leq\tau+\frac{(1+\eta)^2}{2-\nu}
    \leq\frac32.
    \label{eq:app_auxiliary_norm}
\end{equation}
Combining Eqs.~\eqref{eq:app_energy_factorization}--\eqref{eq:app_auxiliary_norm} gives
\begin{equation}
    (\mathbf u+\mathbf r)(\mathbf u+\mathbf r)^\top
    =\mathbf L\boldsymbol\xi\boldsymbol\xi^\top\mathbf L^\top
    \preceq\frac32(\mathbf I-\mathbf M\mathbf M^\top).
    \label{eq:app_joint_energy_bound}
\end{equation}
Thus, the block matrix
$[\mathbf M\;\;\sqrt{2/3}(\mathbf u+\mathbf r)]$
has operator norm at most one.
Restoring the time index, apply this contraction to the stacked recurrent state and gated value to obtain
\begin{equation}
    \begin{aligned}
        \|\mathbf S_t\|_F^2
        &=\left\|
        \begin{bmatrix}
            \mathbf M_t&\sqrt{2/3}(\mathbf u_t+\mathbf r_t)
        \end{bmatrix}
        \begin{bmatrix}
            \mathbf S_{t-1}\\
            \sqrt{3/2}\,\mathbf z_t^\top
        \end{bmatrix}
        \right\|_F^2\\
        &\leq\|\mathbf S_{t-1}\|_F^2+\frac32\|\mathbf z_t\|_2^2.
    \end{aligned}
\end{equation}
Since $\mathbf z_t=\mathbf w_t\odot\mathbf v_t$ and the write gate has entries in $(0,1)$,
$\|\mathbf z_t\|_2\leq\|\mathbf v_t\|_2\leq V$.
Summing the preceding inequality over tokens and using $\mathbf S_0=\mathbf0$ gives
\vspace{-0.1in}
\begin{equation}
    \begin{aligned}
        \|\mathbf S_T\|_F^2
        &\leq\frac32\sum_{t=1}^{T}\|\mathbf z_t\|_2^2\\
        &\leq\frac32\sum_{t=1}^{T}\|\mathbf v_t\|_2^2
        \leq\frac32TV^2.
    \end{aligned}
\end{equation}
\vspace{-0.1in}
Taking square roots proves Eq.~\eqref{eq:stable_state_norm_bound}.
\end{proof}

\subsection{Role of Decay Regulation}
\label{app:decay_regulation_role}

Removing $\sqrt{1-\mathbf d_t^2}$ from the residual leaves the transition unchanged and makes the total write equal to $\mathbf k_t\mathbf z_t^\top$.
Non-expansive transitions are therefore preserved, but the resulting recurrence need not admit an $O(\sqrt T)$ norm bound with a constant independent of the gate values and sequence length.
To see this, consider a scalar invariant direction with constant $k=1$, $b=\varepsilon^2$, $d=1-\varepsilon^2$, and $z=1$, starting from $S_0=0$.
These choices can be embedded along one coordinate of any key dimension; for example, the gated value $z=1$ is realized by $w=1/2$ and $v=2$.
For $0<\varepsilon<1$, the unscaled residual gives
\begin{equation}
    \begin{gathered}
        S_t=mS_{t-1}+1,\qquad m=(1-\varepsilon^2)^2<1,\\
        S_T=\sum_{j=0}^{T-1}m^j.
    \end{gathered}
\end{equation}
For any $T\geq4$, choose $\varepsilon=1/T$.
Bernoulli's inequality gives
$m^{T-1}=(1-T^{-2})^{2(T-1)}\geq1-2(T-1)/T^2\geq1/2$,
and hence $S_T\geq T/2$.
Thus, the values remain bounded by $V=2$, while
$\|\mathbf S_T\|_F\geq T/2$.
For any constant $C$ independent of the gate values and sequence length, this family violates
$\|\mathbf S_T\|_F\leq CV\sqrt T$
for sufficiently large $T$.
The growth comes from repeated writes under weak decay, rather than amplification through the transition products analyzed in Section~\ref{sec:theoretical_analysis}.
This separates the control of write accumulation provided by decay regulation from the non-expansive propagation provided by symmetric erasure.

\subsection{A Bound under Uniform Strict Decay}
\label{app:uniform_decay_bound}

If, in addition, $\|\mathbf d_t\|_\infty\leq\rho<1$ and
$\|\mathbf v_t\|_2\leq V$ for all tokens, then
$\|\mathbf z_t\|_2\leq V$.
Proposition~\ref{prop:nonexpansive_propagation} and Eq.~\eqref{eq:app_write_norm} therefore imply
\vspace{-0.1in}
\begin{equation}
    \begin{aligned}
        \|\mathbf S_T\|_F
        &\leq\rho^T\|\mathbf S_0\|_F
        +V\sum_{t=1}^{T}\rho^{T-t}\\
        &=\rho^T\|\mathbf S_0\|_F
        +V\frac{1-\rho^T}{1-\rho}
        \leq\|\mathbf S_0\|_F+\frac{V}{1-\rho}.
    \end{aligned}
\end{equation}
This bound is independent of sequence length under the additional uniform decay condition.
The $O(\sqrt T)$ norm bound in Proposition~\ref{prop:write_energy_bound} does not require a uniform gap between the decay factors and one.

\end{document}